\documentclass[11pt]{article}

\usepackage[]{acl}
\usepackage{algorithm}
\usepackage{subcaption}
\usepackage{graphicx}
\usepackage{amssymb}
\usepackage{amsthm}
\usepackage{hyperref}
\usepackage{booktabs}
\usepackage{url}
\usepackage{xcolor}
\usepackage[table]{xcolor}
\usepackage{listings}
\usepackage{float}
\usepackage{amsmath}
\usepackage{times}
\usepackage{latexsym}
\usepackage{tcolorbox}

\usepackage[T1]{fontenc}

\usepackage[utf8]{inputenc}
\usepackage{multirow}
\usepackage{microtype}

\usepackage{inconsolata}

\usepackage{graphicx}

\title{When Do Supervised UQ Ensembles Improve LLM Hallucination Detection? A Robustness Study}

\author{ Mohit Singh Chauhan\thanks{Correspondence: \texttt{mohitsingh.chauhan@cvshealth.com}}  \qquad
Vipin Gyanchandani \qquad
Dylan Bouchard \qquad
\\ 
CVS Health\textsuperscript{\textregistered}, Wellesley, MA, USA 
}

\begin{document}

\maketitle

\begin{abstract}
Uncertainty quantification (UQ) methods are widely used for hallucination detection in large language models (LLMs) in closed-book settings where ground-truth evidence is unavailable at inference time. Prior work has proposed combining UQ signals via learned ensembles, but empirical investigations into the robustness of these ensembles are limited. We study a supervised ensembling framework that trains a classifier over heterogeneous UQ-based scorer outputs on a small, domain-specific dataset of labeled LLM responses, then applies it to out-of-sample hallucination classification without retrieval, tools, or reference documents. Across four LLMs, nine datasets, and three generation regimes (short-form QA, long-form generation, and code generation), we provide a systematic robustness analysis along three axes: sample efficiency, in-domain dataset transfer, and generation regime dependence. We find that supervised ensembles outperform the best individual scorer in 30 of 32 settings, with gains realized from as few as 100 labeled instances. Ensembles retain most of their advantage in cases of in-domain transfer under distribution shift, outperforming the best non-ensemble scorer in 23 of 28 transfer settings. Sampling-based black-box ensembles are nearly as effective as full ensembles, while single-generation white-box ensembles offer limited benefit.
\end{abstract}

\section{Introduction}
Large language models (LLMs) are increasingly used as closed-book generators in settings where retrieval is unavailable, including internal knowledge assistants, clinical and policy summarization workflows, customer support composition, and code generation in proprietary repositories. In these deployments, hallucinations are a primary failure mode, as models generate fluent but incorrect statements, fabricated citations, or incorrect code behaviors. Detecting hallucinations at inference time is difficult because the system often lacks a trusted external reference and because hallucination rates vary sharply by domain, prompt distribution, and base LLM, making reliable closed-book detection a practical bottleneck for safe deployment.

Uncertainty quantification (UQ) provides a rich set of signals for hallucination detection, and prior work has proposed black-box methods (e.g., self-consistency across samples or perturbations), white-box methods (e.g., likelihood- or token-probability-derived measures), and reflexive or self-judge methods (e.g., model-generated assessments of its own correctness). These signals are attractive because they can be computed without retrieval, often using only model outputs and, when available, token-level probabilities. In practice, however, no single signal is uniformly superior across generation regimes (e.g. natural language vs. code), and the same scorer can behave differently across domains (e.g. math vs. factual QA) and LLMs, making zero-shot thresholds brittle.

A natural response is to combine multiple UQ scorers via supervised ensembling. Prior work has demonstrated that even simple weighted-average ensembles of black-box, white-box, and judge-based scorers consistently outperform individual components in controlled, in-domain short-form question-answering settings \citep{bouchard2025uncertainty}. However, that work evaluated ensembles only with a fixed combination strategy (linear weighting), only in-distribution, and only on short-form outputs. The robustness of supervised UQ ensembles across realistic deployment conditions, where labeled data may be scarce, test distributions may drift from training, and generation formats vary widely, remains an open question.

In this work, we study a supervised ensembling framework that trains a classifier over a heterogeneous collection of black-box, white-box, and reflexive UQ scorer outputs on a domain-specific dataset of labeled LLM responses, applied to out-of-sample hallucination classification in a closed-book setting. Our primary contribution is a systematic robustness study along three deployment-critical axes: (1) sample efficiency, (2) in-domain transfer under distribution shift, and (3) generation regime dependence across short-form QA, long-form generation, and code generation. Across four LLMs, nine datasets, and three generation regimes, we find that supervised ensembles outperform the best individual scorer in 30 of 32 settings by AUROC and 29 of 32 by calibration (ECE), with gains realized from as few as 100 labeled instances. In cases of in-domain distributional shift, ensembles retain most of their in-distribution advantage, outperforming the best individual scorer in 23 of 28 transfer settings. Black-box-only ensembles are nearly as effective as full ensembles, while white-box-only ensembles offer limited benefit. Among combination strategies, logistic regression offers the best overall balance of performance and stability across regimes.

\section{Related Work}

\paragraph{Uncertainty Quantification for LLMs.}
A variety of UQ methods have been proposed for hallucination detection in LLM outputs \citep{huang2023surveyhallucinationlargelanguage, shorinwa2024surveyuncertaintyquantificationlarge}. 
These methods vary along two dimensions: access requirements (black-box, requiring only text outputs, vs. white-box, requiring token probabilities) and mechanism. Sampling-based consistency methods generate multiple responses to the same prompt and measure consistency via exact match \citep{cole2023selectivelyansweringambiguousquestions}, lexical similarity \citep{kuhn2023semanticuncertaintylinguisticinvariances}, embedding similarity \citep{manakul2023selfcheckgptzeroresourceblackboxhallucination, zhang2020bertscoreevaluatingtextgeneration, shorinwa2024surveyuncertaintyquantificationlarge}, or NLI-based semantic equivalence \citep{chen2023quantifyinguncertaintyanswerslanguage, lin2024generatingconfidenceuncertaintyquantification, Farquhar2024}. These are typically black-box but can incorporate token probabilities as well \citep{qiu2024semanticdensityuncertaintyquantification, kuhn2023semanticuncertaintylinguisticinvariances}. White-box methods aggregate token probabilities into response-level scores through measures such as sequence probability, perplexity, entropy, and probability margins \citep{malinin2021uncertaintyestimationautoregressivestructured, manakul2023selfcheckgptzeroresourceblackboxhallucination, fadeeva2024factcheckingoutputlargelanguage, farr2024redctsystemsdesignmethodology}. Reflexive methods prompt the generating LLM or an external judge to self-evaluate correctness \citep{kadavath2022languagemodelsmostlyknow, chen2023quantifyinguncertaintyanswerslanguage, xiong2024llmsexpressuncertaintyempirical, tian2023justaskcalibrationstrategies}. These scorer families form the individual components of the ensembles we study.

\paragraph{UQ Beyond Short-Form Question Answering.}
Most UQ methods have been developed and evaluated on short-form question answering, and short-form UQ has been shown to generalize poorly to long-form outputs \citep{bakman2025reconsideringllmuncertaintyestimation, vashurin2025uncertaintylinelengthinvariantestimationuncertainty}. Fine-grained methods address this by decomposing responses into sentences or claims and scoring each unit via entailment against sampled responses \citep{zhang2024luqlongtextuncertaintyquantification}, graph centrality over claim-response entailment graphs \citep{jiang2024graphbaseduncertaintymetricslongform}, or question-generation pipelines \citep{Farquhar2024}. For code generation, early studies have investigated token-probability calibration for generated code \citep{spiess2024calibrationcorrectnesslanguagemodels}, symbolic clustering methods \citep{sharma2025assessingcorrectnessllmbasedcode}, and functional equivalence variants of semantic entropy \citep{bouchard2026functionalentropy}. However, no prior work has evaluated the effectiveness of supervised ensembles over these regime-specific scorer families.

\paragraph{Ensemble Approaches.}
For unsupervised ensembles, \citet{chen2023quantifyinguncertaintyanswerslanguage} propose BSDetector, a two-component ensemble that computes a weighted average of observed consistency (combining exact match and NLI scores) and self-reflection certainty, and \citet{verga2024replacingjudgesjuriesevaluating} propose a Panel of LLM evaluators that aggregates judgments from multiple smaller LLMs rather than a single large judge. For supervised ensembles, \citet{bouchard2025uncertainty} tune a weighted average over combinations of black-box, white-box, and judge-based scorers and demonstrate consistent gains over individual components on short-form benchmarks. \citet{bakman2025reconsideringllmuncertaintyestimation} ensemble short-form UQ scorers and find ensembles consistently outperform the best individual short-form method, even with small training datasets.  Our work extends supervised ensemble UQ with a dedicated robustness analysis: we compare combination strategies beyond weighted averaging, evaluate generalization under distribution shift, and study ensemble effectiveness across short-form, long-form, and code generation regimes, the latter two requiring materially different scorer families.

\section{Methods}

\subsection{Problem Formulation}

We frame hallucination detection as binary classification over UQ scorer outputs. Given a prompt $x$ and a generated response $y$, let $\mathbf{s}(y) = (s_1(y), \ldots, s_K(y)) \in [0,1]^K$ denote a vector of $K$ confidence scores produced by a collection of UQ-based scorers, where each $s_k$ maps a response to a scalar confidence in $[0,1]$ with higher values indicating greater confidence in correctness. The ground-truth label $h(y) \in \{0, 1\}$ indicates whether $y$ contains a hallucination ($h=1$) or not ($h=0$), determined by comparison to a reference available only offline. Our goal is to learn a function $f: [0,1]^K \to [0,1]$ that maps the scorer vector to a single confidence score that separates hallucinated from correct responses. 

For long-form generation, we operate at the claim level rather than the response level. Each claim $c$ extracted from a response receives its own scorer vector $\mathbf{s}(c) \in [0,1]^K$, and the ensemble classifies claims individually, where $h(c) \in \{0,1\}$ indicates whether the claim is supported by a reference text available only offline.

\subsection{UQ Scorer Families}
\label{sec:scorers}

The ensemble operates over confidence scores drawn from four families of UQ methods.  Table~\ref{tab:scorer_summary} summarizes all scorers by family, access requirements, and applicable generation regimes, with formal definitions and implementation details provided in Appendix~\ref{sec:scorer_details}. We describe each family below.

\begin{table}[t]
\centering
\tiny
\begin{tabular}{llccc}
\toprule
\textbf{Family} & \textbf{Access} & \textbf{Short} & \textbf{Long} & \textbf{Code} \\
\midrule
\multicolumn{5}{l}{\textit{Single-generation white-box}} \\
\qquad Sequence probability & White-box & \checkmark & & \checkmark \\
\qquad Norm. sequence probability & White-box & \checkmark & & \checkmark \\
\qquad Min token probability & White-box & \checkmark & & \checkmark \\
\qquad Probability margin & White-box & \checkmark & & \checkmark \\
\qquad Mean token entropy & White-box & \checkmark & & \checkmark \\
\qquad Max token entropy & White-box & \checkmark & & \checkmark \\

\multicolumn{5}{l}{\textit{Consistency-based black-box}} \\
\qquad Exact match rate & Black-box & \checkmark & &  \\
\qquad Non-contradiction probability & Black-box & \checkmark & &  \\
\qquad BERTScore consistency & Black-box & \checkmark & &  \\
\qquad Semantic entropy & Black-box & \checkmark & &  \\
\qquad Cosine similarity & Black-box & \checkmark & & \checkmark \\
\qquad Functional entropy & Black-box &  & & \checkmark \\
\qquad Equivalence rate & Black-box & & & \checkmark \\
\qquad CodeBLEU consistency & Black-box & & & \checkmark \\

\multicolumn{5}{l}{\textit{Consistency-based white-box}} \\
\qquad CoCoA & White-box & \checkmark & & \checkmark \\
\qquad Monte Carlo probability & White-box & \checkmark & &  \checkmark\\
\qquad WB semantic entropy & White-box & \checkmark & & \checkmark \\
\qquad Semantic density & White-box & \checkmark & &   \\

\multicolumn{5}{l}{\textit{Reflexive}} \\
\qquad Verbalized confidence & Black-box & \checkmark & \checkmark & \checkmark \\
\qquad P(True) & White-box & \checkmark & \checkmark & \checkmark \\
\multicolumn{5}{l}{\textit{Graph-based (claim-level)}} \\
\qquad Degree Centrality & Black-box & & \checkmark & \\
\qquad Betweenness Centrality & Black-box & & \checkmark & \\
\qquad Closeness Centrality & Black-box & & \checkmark & \\
\qquad Harmonic Centrality & Black-box & & \checkmark & \\
\qquad Laplacian Centrality & Black-box & & \checkmark & \\
\qquad PageRank & Black-box &  & \checkmark & \\
\bottomrule
\end{tabular}
\caption{Summary of UQ scorers used as ensemble inputs by access and generation regime. ``Access'' indicates whether token probability access is required. Formal definitions are in Appendix~\ref{sec:scorer_details}.}
\label{tab:scorer_summary}
\end{table}

\paragraph{Black-box consistency scorers} generate $m$ candidate responses from the same prompt using stochastic decoding and measure agreement between the original response and candidates. Methods differ in their consistency function: exact match \citep{cole2023selectivelyansweringambiguousquestions}, NLI-based non-contradiction or entailment \citep{chen2023quantifyinguncertaintyanswerslanguage, lin2024generatingconfidenceuncertaintyquantification}, embedding cosine similarity \citep{shorinwa2024surveyuncertaintyquantificationlarge}, BERTScore \citep{zhang2020bertscoreevaluatingtextgeneration}, and semantic entropy via NLI-based clustering \citep{kuhn2023semanticuncertaintylinguisticinvariances, Farquhar2024}. For code generation, we additionally employ code-specific consistency functions including CodeBLEU \citep{ren2020codebleumethodautomaticevaluation} and LLM-based functional equivalence assessment, replacing NLI-based semantic comparison with judgments of whether two code snippets produce identical outputs for all valid inputs.

\paragraph{White-box token-probability scorers} derive confidence from the token-level probabilities produced during generation. We consider length-normalized sequence probability \citep{malinin2021uncertaintyestimationautoregressivestructured}, minimum token probability \citep{manakul2023selfcheckgptzeroresourceblackboxhallucination}, probability margin \citep{farr2024redctsystemsdesignmethodology}, and token-level entropy \citep{scalena2025eagerentropyawaregenerationadaptive}. We also consider hybrid methods that combine token probabilities with sampling-based consistency (e.g., white-box semantic entropy \citep{kuhn2023semanticuncertaintylinguisticinvariances}).

\paragraph{Reflexive (judge-based) scorers} prompt the generating LLM or an external LLM to evaluate correctness of a question-response pair. We consider verbalized confidence \citep{tian2023justaskcalibrationstrategies, xiong2024llmsexpressuncertaintyempirical} and P(True) \citep{kadavath2022languagemodelsmostlyknow}. For short-form and code generation, these scorers evaluate the full response. For long-form generation, they are applied at the claim level, scoring each extracted claim individually.

\paragraph{Claim-level scorers (long-form only)} decompose responses into claims and score each claim individually, producing the claim-level confidence scores over which the ensemble operates. Following \citet{jiang2024graphbaseduncertaintymetricslongform}, we employ graph-based scorers, which construct claim-response entailment graphs and use graph centrality metrics to measure uncertainty at the claim level.

\subsection{Ensembling Strategies}
\label{sec:combination}

Given the scorer vector $\mathbf{s} \in [0,1]^K$ and binary labels $h \in \{0,1\}$ for a training set of $n$ labeled instances, we train a classifier $f$ to predict hallucinations from scorer outputs. We compare four combination strategies of varying complexity: (1) logistic regression with $\ell_2$ regularization, (2) random forest, (3) gradient boosted trees, and constrained weighted average \citep{bouchard2025uncertainty}. For all strategies, hyperparameters are selected via 5-fold cross-validation on the tuning set. See Appendix~\ref{sec:hyperparameters} for hyperparameter details.

\section{Experiments}

\subsection{Setup}

We evaluate four LLMs spanning two providers and two capability tiers: Gemini-2.5-Flash, Gemini-2.5-Pro \citep{gemini_doc}, GPT-4o, and GPT-4o-mini \citep{OpenAI_doc}. Experiments are organized across four core domains (Math, Factual QA, Code, and Long-form) and one standalone reading comprehension task. For short-form evaluation, we consider: (1) \textbf{Math reasoning}, consisting of OpenR1-Math \citep{huggingface_open_r1_0_1_0_dev0} and BigMath \citep{albalak2025bigmathlargescalehighqualitymath} (1{,}000 questions each); (2) \textbf{Factual QA}, consisting of HotpotQA \citep{yang2018hotpotqadatasetdiverseexplainable} and SimpleQA \citep{wei2024measuringshortformfactualitylarge} (1{,}000 questions each); and (3) \textbf{Reading Comprehension}, using the DROP dataset \citep{dua2019dropreadingcomprehensionbenchmark} (1{,}000 questions). For code generation, we use two subsets of LiveCodeBench \citep{jain2024livecodebenchholisticcontaminationfree}: a Leetcode-derived callable subset (442 problems) and an AtCoder/CodeForces I/O subset (610 problems), both requiring Python generation. For long-form QA, we construct two datasets following the FactScore \citep{min2023factscorefinegrainedatomicevaluation} protocol: world's largest rivers (500 questions) and edible mushrooms (84 questions), with responses averaging approximately 32 claims each, yielding roughly 16{,}000 and 2{,}700 claim-level instances per LLM, respectively (see Appendix~\ref{sec:longform_datasets} for more details). 

For short-form questions, hallucination labels are obtained by comparing LLM responses to reference answers using an LLM-based grading procedure. For code generation, labels are determined by execution against test cases (pass@1). For long-form QA, responses are decomposed into claims and each claim is graded against the corresponding Wikipedia article using the FactScore protocol. Gemini-2.5-Flash is used as the grading model for short-form and long-form questions, as well as for claim decomposition in the long-form setting, chosen for its strong performance at low cost.  For all sampling-based scorers, we generate 10 sampled responses per prompt across all regimes.

To assess the reliability of LLM-based grading, two human annotators independently labeled a stratified sample of 400 short-form responses (20 correct and 20 incorrect per dataset per generator LLM, spanning all five short-form datasets and two generators: GPT-4o and Gemini-2.5-Flash). Annotators compared each generated answer against the reference answer without access to the grader's label. Human-human agreement was 97.5\% (Cohen's $\kappa = 0.95$), and LLM-human agreement was 98.8\% ($\kappa = 0.97$) and 96.2\% ($\kappa = 0.93$) for the two annotators, confirming that grading noise is no larger than inherent human disagreement. Agreement rates were comparable across both generator LLMs ($\kappa = 0.95$ for both), providing no evidence that the grader favors its own responses. Full results broken down by dataset and generator are provided in Appendix~\ref{sec:grading_validation}.

All experiments share a common splitting procedure: for each domain with paired datasets, we generate 25 random stratified 70/30 splits applied simultaneously to both datasets, yielding a 70\% training fold and 30\% test fold for each.\footnote{DROP is included as a standalone reading comprehension task for which we do not have a companion dataset. The same 25-split procedure is applied for in-distribution evaluation; DROP is excluded from the in-domain transfer analysis.} The three analyses described below (in-distribution performance, in-domain transfer, and access-constrained ablations) all operate over these same 25 splits. Table~\ref{tab:llm_accuracy} reports LLM accuracy rates for each dataset. Complete AUROC and ECE results, broken down by scorer and ensemble for all LLM-dataset combinations, are provided in Tables~\ref{tab:auroc_results_flash}--\ref{tab:ece_results_mini}.

\begin{figure*}[t]
    \centering
    
    \begin{subfigure}[b]{0.95\textwidth}
        \includegraphics[width=\textwidth]{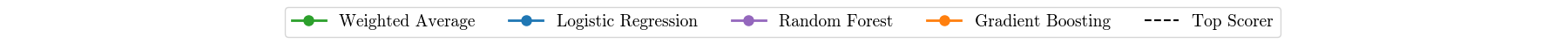}
    \end{subfigure}
    \vspace{2mm}

    \begin{subfigure}[b]{0.95\textwidth}
        \includegraphics[width=\textwidth]{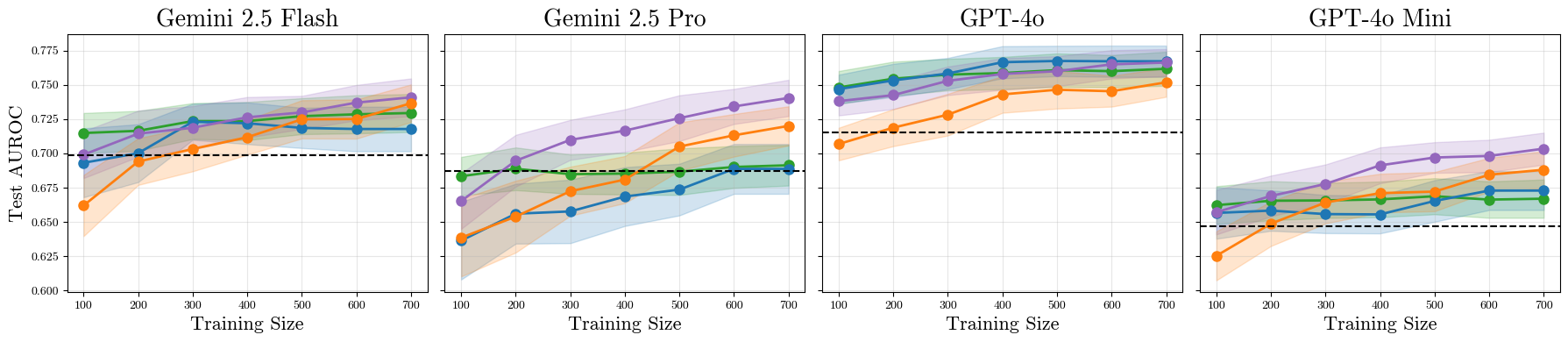}
        \caption{Short-form QA (DROP)}
        \label{fig:drop_learning}
    \end{subfigure}
    \vspace{2mm}

    \begin{subfigure}[b]{0.95\textwidth}
        \includegraphics[width=\textwidth]{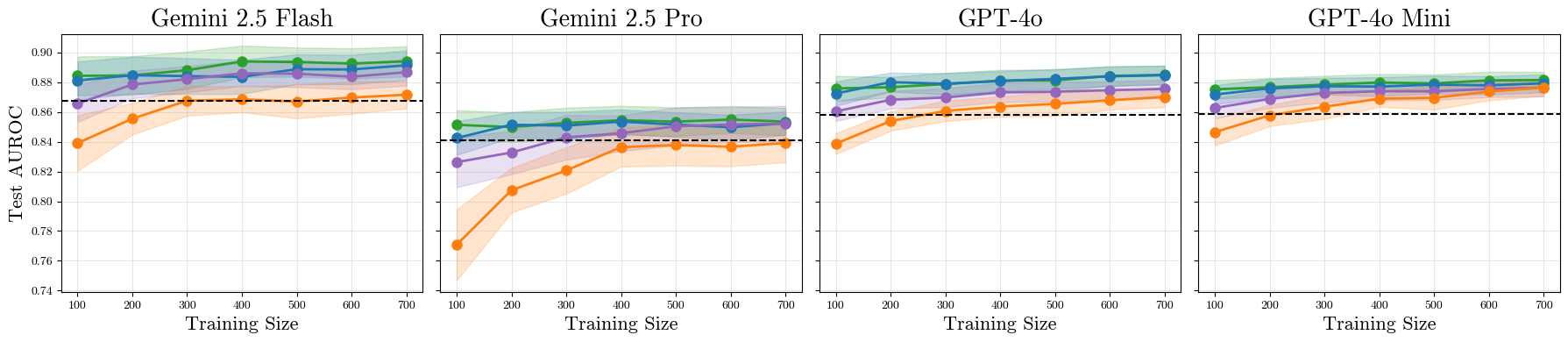}
        \caption{Code Generation (LiveCodeBench)}
        \label{fig:livecodebench_learning}
    \end{subfigure}
    \vspace{2mm}

    \begin{subfigure}[b]{0.95\textwidth}
        \includegraphics[width=\textwidth]{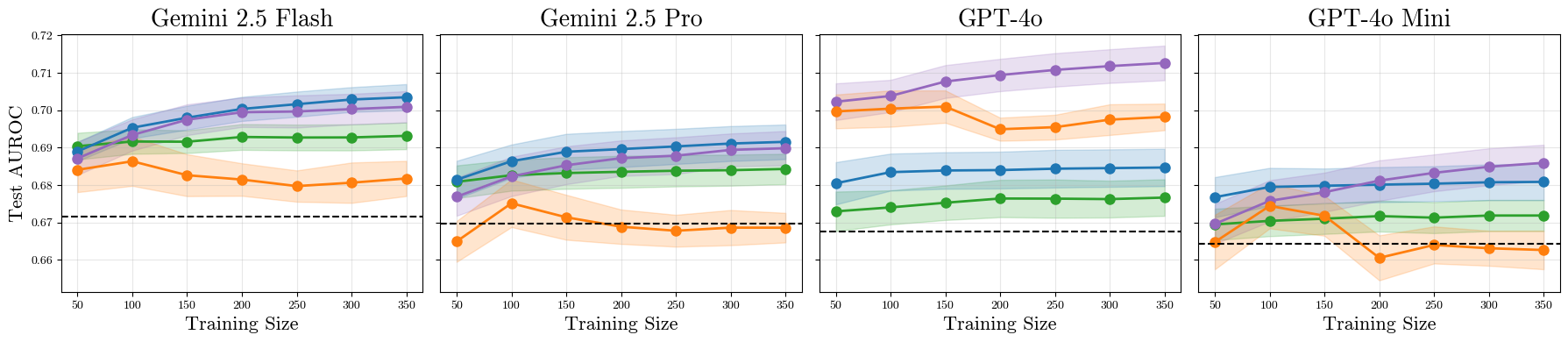}
        \caption{Long-form QA (FactScore-Rivers)}
        \label{fig:rivers_learning}
    \end{subfigure}
\caption{Ensemble AUROC as a function of training sample size for one representative dataset per generation regime. Lines show the four combination strategies with 95\% CIs over 25 splits. The dashed line is the best individual scorer, selected on the test set and therefore an optimistic baseline unavailable in practice. Training sizes (in number of responses) range from $0.1N$ to $0.7N$, where $N$ is the dataset size. Code generation combines both LiveCodeBench subsets. Results for the remaining five datasets are in Figure~\ref{fig:additional_learning_curves}.}
    \label{fig:learning_curves}
\end{figure*}

\subsection{In-Distribution Performance}
For each of the 25 splits, we train ensembles on subsamples of the training fold at sizes $0.1N, 0.2N, \ldots, 0.7N$, where $N$ is the dataset size and $0.7N$ corresponds to the full training fold. For each split and sample size, we draw a single subsample, train the ensemble, and evaluate on the test fold, reporting mean AUROC and 95\% confidence intervals across the 25 splits. For code generation, the two LiveCodeBench subsets are combined into a single pool. The best individual scorer, defined as the scorer achieving the highest AUROC on the test set, serves as a fixed reference baseline. Note that this baseline is optimistic, as it requires test-set access and is unavailable in practice, so ensemble gains relative to a realistic validation-selected scorer would be at least as large.
 
When using the full training sample ($0.7N$), the best ensemble outperforms the best individual scorer in 30 of 32 LLM-dataset settings by AUROC, including every code generation and long-form setting. The two exceptions are BigMath for Gemini-2.5-Flash (best ensemble 0.85 vs.\ scorer 0.86) and SimpleQA for GPT-4o-mini (both 0.78). Results are also consistent under the Prediction Rejection Ratio (PRR) \citep{Vashurin_2025}, with the best ensemble achieving the highest PRR in 30 of 32 settings. We next examine how quickly these gains emerge as a function of training set size. Figures~\ref{fig:openr1_learning}--\ref{fig:mushrooms_learning} show ensemble AUROC as a function of training sample size across all domains. Each panel displays the four combination strategies and the best individual scorer baseline.

\paragraph{Short-form (Figures~\ref{fig:drop_learning}, \ref{fig:openr1_learning}--\ref{fig:hotpotqa_learning}).} At 700 training samples, random forest and logistic regression are the strongest combination strategies in most settings, though neither dominates uniformly. Random forest achieves the highest AUROC in settings with clear ensemble gains (e.g., 0.90 on OpenR1-Math for both Gemini models, 0.74 on DROP for Gemini-2.5-Flash and GPT-4o), while logistic regression leads or ties on datasets where the best individual scorer is already strong (e.g., HotpotQA for Gemini-2.5-Flash, SimpleQA for GPT-4o-mini). Gradient boosting is competitive at 700 samples in some settings but lags at small sample sizes. Weighted average is competitive on datasets with strong baselines (e.g., SimpleQA, HotpotQA) but falls substantially behind on OpenR1-Math, where it plateaus around 0.80--0.86 compared to 0.90 for random forest.
 
Convergence speed varies by strategy. The weighted average stabilizes early, often by 100-200 samples. Logistic regression converges by 200-300 samples in most settings, though it continues improving through 400-500 on some datasets (e.g., OpenR1-Math). Random forest and gradient boosting continue improving with larger samples, with random forest typically reaching its peak earlier.

\paragraph{Code generation (Figure~\ref{fig:livecodebench_learning}).} The combined LiveCodeBench dataset yields consistent ensemble gains across all four LLMs: Gemini-2.5-Flash (0.89 vs.\ 0.87), Gemini-2.5-Pro (0.85 vs.\ 0.84), GPT-4o (0.88 vs.\ 0.86), and GPT-4o-mini (0.88 vs.\ 0.86). The weighted average, logistic regression, and random forest converge to similar AUROC in all four settings, with gradient boosting trailing by 0.01--0.02 for three of the four LLMs and matching the others for GPT-4o-mini. Convergence is fast, with weighted average stabilizing by 100–200 samples and logistic regression and random forest by 200–400.

\paragraph{Long-form QA (Figures~\ref{fig:rivers_learning}, \ref{fig:mushrooms_learning}).} The long-form setting operates at the claim level, with responses averaging approximately 32 claims each. The best ensemble outperforms the best individual scorer in all 8 settings. On the Mushrooms dataset, logistic regression and the weighted average lead for Gemini models and GPT-4o-mini, while all four strategies perform similarly for GPT-4o. On the Rivers dataset, random forest is competitive with or outperforms logistic regression for all LLMs, and substantially outperforms the weighted average for GPT-4o (0.713 vs.\ 0.677) and GPT-4o-mini (0.686 vs.\ 0.672). Gradient boosting underperforms in the long-form setting, degrading with increasing training data for Gemini-2.5-Pro on Rivers (dropping from 0.675 at $0.2N$ to 0.669 at $0.7N$) and GPT-4o-mini (dropping from 0.674 to 0.663). The simpler strategies converge rapidly, typically by $0.1N$--$0.2N$.

\paragraph{Calibration.} Ensembling also improves calibration considerably (Tables~\ref{tab:ece_results_flash}--\ref{tab:ece_results_mini}). In particular, across all 32 LLM-dataset settings, the best ensemble achieves the lowest ECE among all individual scorers and ensemble variants in 29 of 32 cases, with ECE never exceeding 0.06.

\begin{figure*}[t]
    \centering
    
    \begin{subfigure}[b]{0.55\textwidth}
        \includegraphics[width=\textwidth]{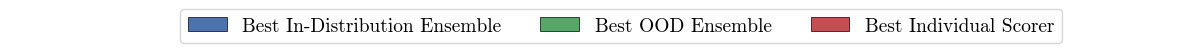}
    \end{subfigure}
    \vspace{1mm}

    \centering
    \begin{subfigure}[b]{0.48\textwidth}
        \includegraphics[width=\textwidth]{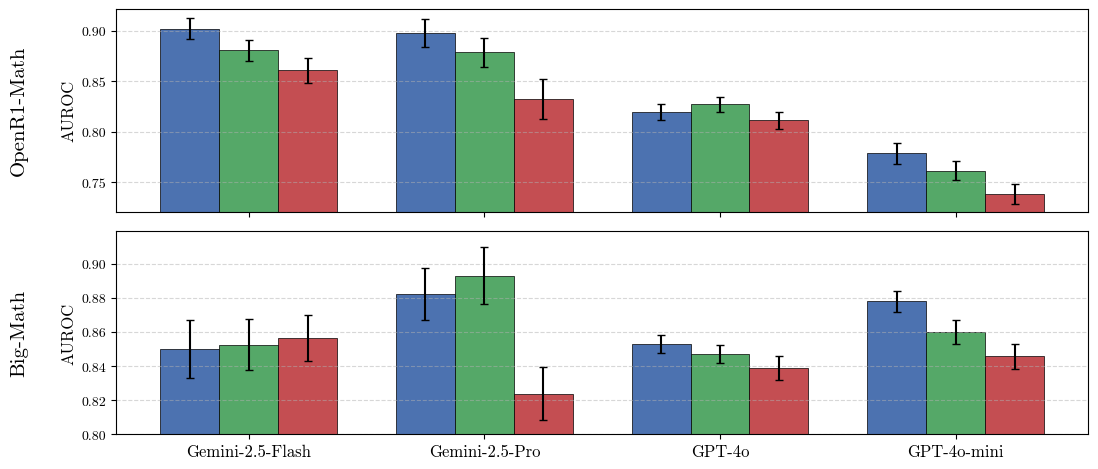}
        \caption{Short-Form Math}
        \label{fig:math_ood}
    \end{subfigure}
    \hfill
    \begin{subfigure}[b]{0.48\textwidth}
        \includegraphics[width=\textwidth]{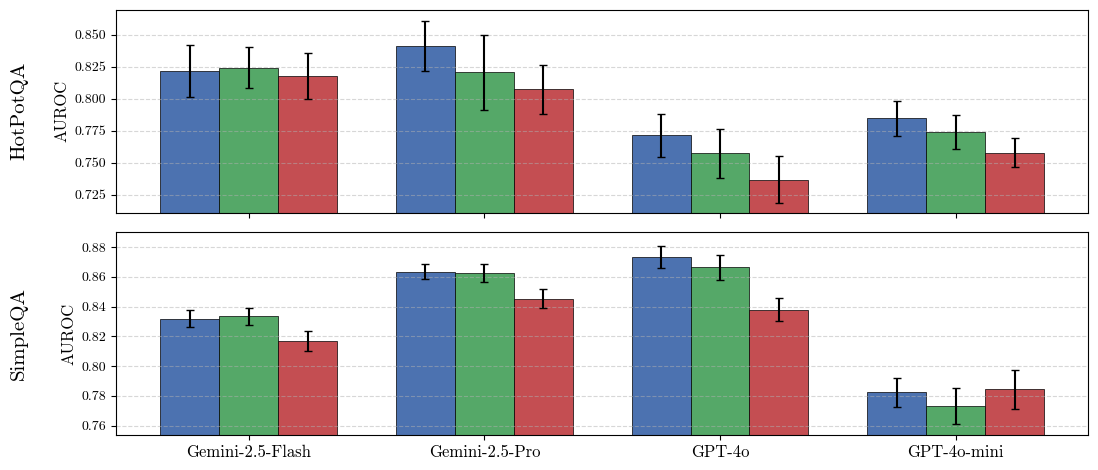}
        \caption{Short-Form Factual QA}
        \label{fig:qa_ood}
    \end{subfigure}
    
    \vspace{0.5em}

    \begin{subfigure}[b]{0.48\textwidth}
        \includegraphics[width=\textwidth]{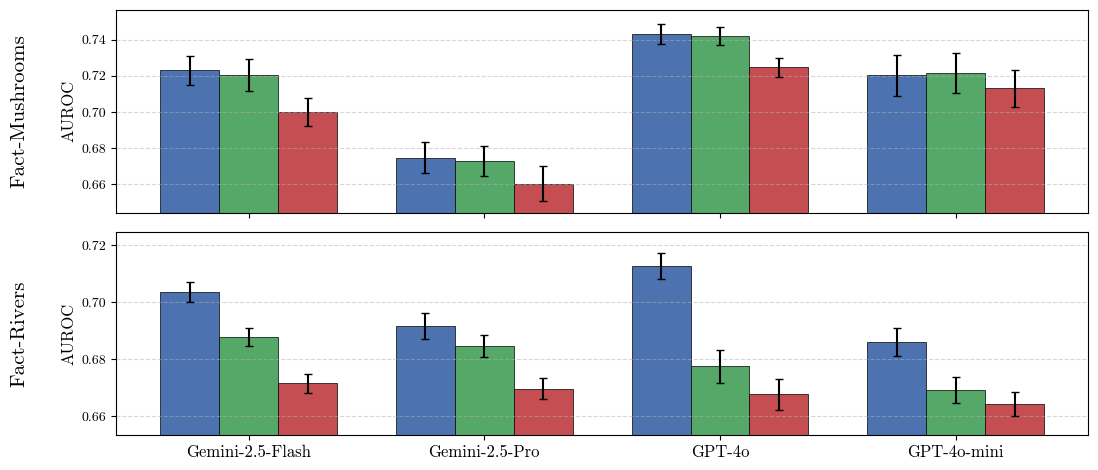}
        \caption{Long-Form QA}
        \label{fig:longform_ood}
    \end{subfigure}
 \hfill
        \begin{subfigure}[b]{0.48\textwidth}
        \includegraphics[width=\textwidth]{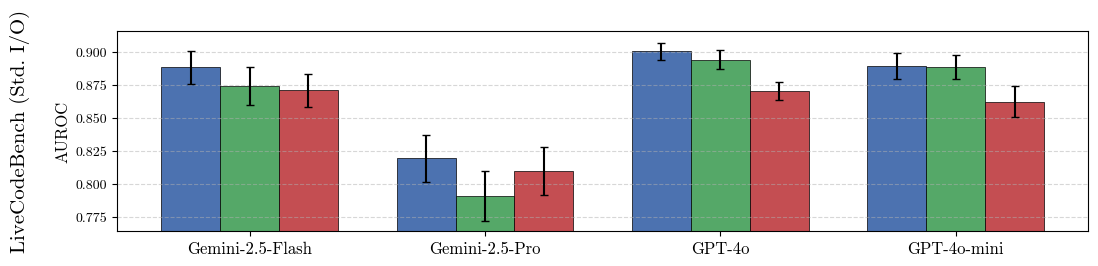}
        \caption{Python Code Generation}
        \label{fig:codegen_ood}
    \end{subfigure}

\caption{In-domain transfer AUROC across domains. Bars compare the best individual scorer (selected on the test set; an optimistic baseline), best in-distribution ensemble, and best out-of-distribution (OOD) ensemble (trained on the companion dataset within the same domain). Combination strategy is selected independently per condition. Bars show mean AUROC with 95\% CIs over 25 splits. Code generation evaluates transfer in one direction only (I/O $\to$ Callable) due to small sample size of Callable subset.}
    \label{fig:ood_comparison}
\end{figure*}

\begin{table*}[t]
\centering
\scriptsize
\begin{tabular}{llccccc}
\toprule
\textbf{LLM} & \textbf{Type} & Big-Math & OpenR1 & DROP & HotpotQA & SimpleQA \\
\midrule
\multirow{6}{*}{Gemini 2.5 Flash} & Top WB Scorer & $0.744 \pm 0.02$ & $0.697 \pm 0.01$ & $0.686 \pm 0.01$ & $0.769 \pm 0.02$ & $0.611 \pm 0.01$ \\
 & Top WB Ensemble & $0.737 \pm 0.02$ & $0.744 \pm 0.01$ & $0.686 \pm 0.01$ & $0.767 \pm 0.02$ & $0.611 \pm 0.01$ \\
 & Top BB Scorer & $0.796 \pm 0.02$ & $0.688 \pm 0.01$ & $0.673 \pm 0.01$ & $0.828 \pm 0.01$ & $0.816 \pm 0.01$ \\
 & Top BB Ensemble & $0.836 \pm 0.02$ & $0.785 \pm 0.01$ & $0.689 \pm 0.01$ & $\mathbf{0.857 \pm 0.01}$ & $0.823 \pm 0.01$ \\
 & Top Overall Scorer & $\mathbf{0.857 \pm 0.01}$ & $0.861 \pm 0.01$ & $0.699 \pm 0.01$ & $0.818 \pm 0.02$ & $0.817 \pm 0.01$ \\
 & Top Overall Ensemble & $0.850 \pm 0.02$ & $\mathbf{0.902 \pm 0.01}$ & $\mathbf{0.741 \pm 0.01}$ & $0.822 \pm 0.02$ & $\mathbf{0.832 \pm 0.01}$ \\
\midrule
\multirow{6}{*}{Gemini 2.5 Pro} & Top WB Scorer & $0.722 \pm 0.02$ & $0.617 \pm 0.02$ & $0.687 \pm 0.02$ & $0.736 \pm 0.02$ & $0.542 \pm 0.01$ \\
 & Top WB Ensemble & $0.717 \pm 0.02$ & $0.634 \pm 0.02$ & $0.689 \pm 0.02$ & $0.733 \pm 0.02$ & $0.539 \pm 0.01$ \\
 & Top BB Scorer & $0.775 \pm 0.02$ & $0.832 \pm 0.02$ & $0.651 \pm 0.02$ & $0.766 \pm 0.03$ & $0.845 \pm 0.01$ \\
 & Top BB Ensemble & $0.822 \pm 0.02$ & $0.843 \pm 0.02$ & $0.687 \pm 0.01$ & $0.786 \pm 0.02$ & $0.851 \pm 0.01$ \\
 & Top Overall Scorer & $0.824 \pm 0.02$ & $0.832 \pm 0.02$ & $0.687 \pm 0.02$ & $0.807 \pm 0.02$ & $0.845 \pm 0.01$ \\
 & Top Overall Ensemble & $\mathbf{0.882 \pm 0.02}$ & $\mathbf{0.898 \pm 0.01}$ & $\mathbf{0.740 \pm 0.01}$ & $\mathbf{0.841 \pm 0.02}$ & $\mathbf{0.863 \pm 0.01}$ \\
\midrule
\multirow{6}{*}{GPT-4o} & Top WB Scorer & $0.839 \pm 0.01$ & $0.806 \pm 0.01$ & $0.696 \pm 0.01$ & $0.732 \pm 0.02$ & $0.842 \pm 0.01$ \\
 & Top WB Ensemble & $0.848 \pm 0.01$ & $0.810 \pm 0.01$ & $0.696 \pm 0.01$ & $0.730 \pm 0.02$ & $0.852 \pm 0.01$ \\
 & Top BB Scorer & $0.798 \pm 0.01$ & $0.783 \pm 0.01$ & $0.715 \pm 0.01$ & $0.704 \pm 0.02$ & $0.840 \pm 0.01$ \\
 & Top BB Ensemble & $0.810 \pm 0.00$ & $0.788 \pm 0.01$ & $0.758 \pm 0.01$ & $0.763 \pm 0.02$ & $0.856 \pm 0.01$ \\
 & Top Overall Scorer & $0.839 \pm 0.01$ & $0.811 \pm 0.01$ & $0.715 \pm 0.01$ & $0.737 \pm 0.02$ & $0.838 \pm 0.01$ \\
 & Top Overall Ensemble & $\mathbf{0.853 \pm 0.01}$ & $\mathbf{0.819 \pm 0.01}$ & $\mathbf{0.767 \pm 0.01}$ & $\mathbf{0.771 \pm 0.02}$ & $\mathbf{0.874 \pm 0.01}$ \\
\midrule
\multirow{6}{*}{GPT-4o Mini} & Top WB Scorer & $0.847 \pm 0.01$ & $0.713 \pm 0.01$ & $0.625 \pm 0.01$ & $0.748 \pm 0.01$ & $0.781 \pm 0.01$ \\
 & Top WB Ensemble & $0.858 \pm 0.01$ & $0.754 \pm 0.01$ & $0.632 \pm 0.01$ & $0.754 \pm 0.01$ & $0.783 \pm 0.01$ \\
 & Top BB Scorer & $0.748 \pm 0.01$ & $0.739 \pm 0.01$ & $0.644 \pm 0.02$ & $0.747 \pm 0.02$ & $0.778 \pm 0.01$ \\
 & Top BB Ensemble & $0.800 \pm 0.01$ & $0.746 \pm 0.01$ & $0.678 \pm 0.01$ & $0.775 \pm 0.01$ & $0.778 \pm 0.01$ \\
 & Top Overall Scorer & $0.846 \pm 0.01$ & $0.738 \pm 0.01$ & $0.647 \pm 0.01$ & $0.758 \pm 0.01$ & $\mathbf{0.784 \pm 0.01}$ \\
 & Top Overall Ensemble & $\mathbf{0.878 \pm 0.01}$ & $\mathbf{0.778 \pm 0.01}$ & $\mathbf{0.703 \pm 0.01}$ & $\mathbf{0.785 \pm 0.01}$ & $0.782 \pm 0.01$ \\
\bottomrule
\end{tabular}%
\caption{Comparison of white-box (WB), black-box (BB), and overall uncertainty quantification methods across five short-form datasets and four LLMs. Each row reports the best individual scorer and best ensemble within that access category. AUROC values with 95\% confidence intervals over 25 random train/test splits. Bold indicates the highest AUROC per LLM-dataset pair.}
\label{tab:performance}
\end{table*}

\subsection{In-Domain Transfer}

For each domain, we use the same 25 splits to evaluate in-domain transfer (Figure~\ref{fig:ood_comparison}). Within each split, we train the ensemble on the training fold of one dataset and evaluate on the test folds of both datasets in the domain: the same dataset (in-distribution) and its companion (transfer). For example, in the math domain, an ensemble trained on OpenR1-Math is evaluated on both OpenR1-Math and BigMath test folds, and vice versa. For each evaluation dataset, we compare the best individual scorer, the best in-distribution ensemble, and the best transfer ensemble, with the combination strategy selected independently for the latter two.

In 13 of 28 settings, the transfer ensemble shows no degradation relative to the in-distribution ensemble, and the maximum degradation across all settings is 0.03 AUROC. Across all 28 settings, the transfer ensemble outperforms the best individual scorer in 23 cases, ties in 2, and underperforms in 3. The 3 underperforming cases are BigMath for Gemini-2.5-Flash (transfer 0.85 vs.\ scorer 0.86), SimpleQA for GPT-4o-mini (0.77 vs.\ 0.78), and code generation for Gemini-2.5-Pro (0.79 vs.\ 0.81), all within confidence intervals. Transfer is consistently strong across domains, with degradation of 0.00--0.03 AUROC and the transfer ensemble matching or exceeding the best scorer in the large majority of settings. Long-form QA shows particularly robust transfer, with no degradation on Mushrooms for any LLM and degradation of at most 0.03 on Rivers.

\subsection{Access-Constrained Ablations}
\label{sec:access_constrained}

Using the same 25 splits and the full training fold, we train ensembles under two restricted access conditions on the five short-form datasets: black-box scorers only (no token probabilities) and single-generation white-box scorers only (no sampling). Table~\ref{tab:performance} reports AUROC for the best individual scorer and best ensemble under each condition, alongside the full scorer set.

\paragraph{Full ensemble.} The full ensemble outperforms the best individual scorer in 18 of 20 LLM-dataset combinations. The two exceptions (BigMath for Gemini-2.5-Flash and SimpleQA for GPT-4o-mini) are within confidence intervals. The largest gains appear for Gemini-2.5-Pro (0.04--0.07 AUROC on math datasets).

\paragraph{Black-box only.} The black-box ensemble outperforms the best black-box scorer in 19 of 20 settings. On HotpotQA for Gemini-2.5-Flash, the black-box ensemble (0.857) exceeds even the best full ensemble (0.822).

\paragraph{White-box only.} The white-box ensemble outperforms the best white-box scorer in just 11 of 20 settings. For GPT-4o and GPT-4o-mini, white-box scorers are competitive with or exceed black-box scorers on several datasets (e.g., BigMath, OpenR1 for GPT-4o). 

\section{Discussion}

\paragraph{Ensembling as a default strategy.} Supervised ensembles that combine black-box and white-box scorers outperform the best individual scorer in 30 of 32 settings by AUROC and 29 of 32 by ECE. This consistency across four LLMs, nine datasets, and three generation regimes suggests that ensembling makes a strong default approach whenever labeled data is available. Crucially, since individual scorers require no training, the labeling effort needed to identify the best single scorer already suffices to train an ensemble, making the additional cost of ensembling negligible. This supervised approach also consistently outperforms training-free, fixed-weight combinations of token probabilities and consistency signals such as CoCoA \citep{vashurin2025uncertaintyquantificationllmsminimum}, which is never the top-performing scorer in our framework.

\paragraph{Calibration.} Beyond discrimination, ensembling yields well-calibrated confidence scores, with the best ensemble achieving ECE below 0.06 in every setting and below 0.05 in most (Tables~\ref{tab:ece_results_flash}--\ref{tab:ece_results_mini}). Individual scorers, by contrast, are often poorly calibrated despite reasonable AUROC. This is practically significant, as well-calibrated scores enable threshold-based deployment decisions (e.g., flagging responses below a confidence threshold for human review) without requiring extensive per-dataset threshold tuning.

\paragraph{Combination strategy selection.} The optimal combination strategy varies by generation regime. For short-form and code generation, random forest and logistic regression are the strongest strategies, with random forest offering marginally higher AUROC at the cost of slower convergence. For long-form claim-level detection, logistic regression and the weighted average dominate, while gradient boosting is prone to overfitting and can degrade with increasing training data. In practice, logistic regression offers the best balance, as it is competitive across all regimes, converges quickly (often by 100--200 responses), and avoids the overfitting risks of tree-based methods in low-feature settings.

\paragraph{Sample efficiency.} Ensemble gains are realized at modest sample sizes. Simpler strategies (logistic regression, weighted average) often reach their plateau with as few as 100--200 labeled instances. Tree-based strategies require 300--500 to converge but can achieve higher AUROC. For practitioners with limited labeling budgets, logistic regression provides strong performance; with larger budgets, random forest may offer incremental improvement.

\paragraph{In-domain transfer.} Transfer ensembles retain most of their in-distribution advantage, with mean degradation of only 0.02 AUROC points and transfer ensembles outperforming the best individual scorer in 23 of 28 settings. This suggests cross-dataset deployment viability, as a practitioner can train an ensemble on a labeled source dataset and deploy it to a related target dataset with reasonable confidence. The few underperforming cases are within confidence intervals rather than being clearly attributable to a systematic failure mode.

\paragraph{Access constraints.} Black-box ensembles provide nearly the same benefit as full ensembles (19 of 20 short-form settings), making them a useful default when logprobs are unavailable. The diversity of black-box consistency signals spanning exact match, NLI, embedding similarity, and semantic clustering provides sufficient complementarity for a classifier to exploit. White-box-only ensembles, by contrast, offer limited gains (11 of 20), as these six token-probability features seem to lack the diversity needed for effective combination. The informativeness of white-box scorers also varies across model families, with GPT models producing consistently useful token-probability signals but Gemini models showing cases where white-box scorers are uninformative or misleading (e.g., Gemini-2.5-Pro on SimpleQA). 

\paragraph{Comparison with prior ensembling studies.} Our findings are consistent with \citet{bakman2025reconsideringllmuncertaintyestimation}, who evaluate ensembling of short-form UQ-based scorers and find that linear ensembles outperform the best individual method with as few as 100 calibration samples. We similarly find that simpler strategies plateau by 100--200 labeled instances. One divergence is that their single decision tree is generally uncompetitive, whereas our tree-based strategies (random forest, gradient boosting) are competitive in most settings. A plausible explanation is that bagging and boosting with cross-validated depth reduce the variance that makes individual trees unreliable. Our study complements theirs by providing a dedicated robustness analysis investigating cross-dataset transfer, access constraints, and two additional generation regimes (code generation and claim-level scoring).

\section{Conclusion}

We presented a systematic robustness study of supervised UQ ensembles for LLM hallucination detection across sample efficiency, in-domain distribution shift, and generation regime. Ensembles outperform the best individual scorer in 30 of 32 settings by AUROC and 29 of 32 by calibration, with gains realized at sample sizes as small as 100 labeled instances. Cross-dataset transfer is effective in most settings (23 of 28), with the primary failure mode being scorer signals that shift across distributions. Black-box-only ensembles are nearly as effective as full ensembles, while white-box-only ensembles offer limited benefit. These findings provide actionable guidance: supervised ensembling is a reliable, low-cost default for hallucination detection when even a small labeled dataset is available, and black-box ensembles are a robust fallback when token probabilities are unavailable. 

\section*{Limitations}
\paragraph{Closed-weight model scope.} Our study evaluates four LLMs from two providers (Google and OpenAI), all of which are closed-source. Results may not generalize to open-weight models (e.g., LLaMA, Mistral), which may exhibit different token-probability characteristics, or to models with substantially different architectures. The observed differences between Gemini and GPT models (particularly in white-box scorer informativeness) suggest that model family is an important factor, but we cannot characterize this dimension fully with only two providers. Moreover, open-weight models enable internal-state methods based on hidden representations or attention maps \citep{azaria-mitchell-2023-internal, chen2024inside, chuang-etal-2024-lookback, vazhentsev-etal-2025-unconditional} that are infeasible with closed-weight APIs. Whether ensembles over these internal-state scorers can match the performance of sampling-based ensembles at lower inference cost is an open question for the open-weight setting.

\paragraph{Evaluation scope.} Our evaluation covers specific slices of each generation regime and transfer condition. The long-form evaluation is limited to two narrow factoid-recall domains (rivers and mushrooms) with a shared question template; broader tasks such as open-ended summarization, document drafting, or multi-turn dialogue may exhibit different ensemble behavior, particularly if claim decomposition is less straightforward. Code generation experiments use Python exclusively on competitive programming problems from LiveCodeBench, whereas real-world code generation spans multiple languages, longer codebases, and tasks beyond self-contained function synthesis. Our cross-dataset transfer evaluation considers transfer between dataset pairs within the same domain (e.g., math to math); cross-domain transfer (e.g., training on math and deploying to factual QA) and cross-LLM transfer (e.g., training on GPT-4o outputs and deploying to Gemini) remain unexplored and may exhibit substantially larger degradation.

\section*{Ethical Considerations.}
This work aims to improve the reliability of LLM outputs by detecting hallucinations, which we view as a net positive for safe deployment. However, high-performing hallucination detectors could create a false sense of security if practitioners treat ensemble confidence scores as guarantees of correctness rather than probabilistic estimates. We emphasize that our methods reduce but do not eliminate hallucination risk, and that human oversight remains essential in high-stakes settings such as clinical or legal applications. We use publicly available research benchmarks and cite their original sources; users should obtain the datasets from the official sources and comply with the corresponding licenses and terms of use. We do not redistribute the full benchmark-derived data, and release only synthetic schema-compatible files for code smoke testing. No personally identifiable information was collected or generated. The human annotation study was conducted by the authors; no crowdworkers were employed.


\section*{Disclosure}
\paragraph{Conflict of interest.} MSC is employed by CVS Health\textsuperscript{\textregistered} Corporation and holds stock and/or equity. VG and DB were formerly employed by CVS Health\textsuperscript{\textregistered} Corporation and hold stock and/or equity. No conflicts germane to this work.

\paragraph{Disclaimer.} Prompts are included solely for reproducibility and do not imply endorsement or affiliation. Gemini is a trademark of Google and GPT is a trademark of OpenAI. This is an independent publication and has not been authorized, endorsed, or sponsored by Google or OpenAI.

\paragraph{LLM usage.} The authors used LLMs to assist with editing the manuscript.

\bibliography{refs.bib}

@misc{albalak2025bigmathlargescalehighqualitymath,
      title={Big-Math: A Large-Scale, High-Quality Math Dataset for Reinforcement Learning in Language Models}, 
      author={Alon Albalak and Duy Phung and Nathan Lile and Rafael Rafailov and Kanishk Gandhi and Louis Castricato and Anikait Singh and Chase Blagden and Violet Xiang and Dakota Mahan and Nick Haber},
      year={2025},
      eprint={2502.17387},
      archivePrefix={arXiv},
      primaryClass={cs.LG},
      url={https://arxiv.org/abs/2502.17387}, 
}

@misc{wei2024measuringshortformfactualitylarge,
      title={Measuring short-form factuality in large language models}, 
      author={Jason Wei and Nguyen Karina and Hyung Won Chung and Yunxin Joy Jiao and Spencer Papay and Amelia Glaese and John Schulman and William Fedus},
      year={2024},
      eprint={2411.04368},
      archivePrefix={arXiv},
      primaryClass={cs.CL},
      url={https://arxiv.org/abs/2411.04368}, 
}

@inproceedings{azaria-mitchell-2023-internal,
    title = "The Internal State of an {LLM} Knows When It{'}s Lying",
    author = "Azaria, Amos  and
      Mitchell, Tom",
    editor = "Bouamor, Houda  and
      Pino, Juan  and
      Bali, Kalika",
    booktitle = "Findings of the Association for Computational Linguistics: EMNLP 2023",
    month = dec,
    year = "2023",
    address = "Singapore",
    publisher = "Association for Computational Linguistics",
    url = "https://aclanthology.org/2023.findings-emnlp.68/",
    doi = "10.18653/v1/2023.findings-emnlp.68",
    pages = "967--976"
}

@inproceedings{
chen2024inside,
title={{INSIDE}: {LLM}s' Internal States Retain the Power of Hallucination Detection},
author={Chao Chen and Kai Liu and Ze Chen and Yi Gu and Yue Wu and Mingyuan Tao and Zhihang Fu and Jieping Ye},
booktitle={The Twelfth International Conference on Learning Representations},
year={2024},
url={https://openreview.net/forum?id=Zj12nzlQbz}
}

@inproceedings{chuang-etal-2024-lookback,
    title = "Lookback Lens: Detecting and Mitigating Contextual Hallucinations in Large Language Models Using Only Attention Maps",
    author = "Chuang, Yung-Sung  and
      Qiu, Linlu  and
      Hsieh, Cheng-Yu  and
      Krishna, Ranjay  and
      Kim, Yoon  and
      Glass, James R.",
    editor = "Al-Onaizan, Yaser  and
      Bansal, Mohit  and
      Chen, Yun-Nung",
    booktitle = "Proceedings of the 2024 Conference on Empirical Methods in Natural Language Processing",
    month = nov,
    year = "2024",
    address = "Miami, Florida, USA",
    publisher = "Association for Computational Linguistics",
    url = "https://aclanthology.org/2024.emnlp-main.84/",
    doi = "10.18653/v1/2024.emnlp-main.84",
    pages = "1419--1436"
}

@inproceedings{vazhentsev-etal-2025-unconditional,
    title = "Unconditional Truthfulness: Learning Unconditional Uncertainty of Large Language Models",
    author = "Vazhentsev, Artem  and
      Fadeeva, Ekaterina  and
      Xing, Rui  and
      Kuzmin, Gleb  and
      Lazichny, Ivan  and
      Panchenko, Alexander  and
      Nakov, Preslav  and
      Baldwin, Timothy  and
      Panov, Maxim  and
      Shelmanov, Artem",
    editor = "Christodoulopoulos, Christos  and
      Chakraborty, Tanmoy  and
      Rose, Carolyn  and
      Peng, Violet",
    booktitle = "Proceedings of the 2025 Conference on Empirical Methods in Natural Language Processing",
    month = nov,
    year = "2025",
    address = "Suzhou, China",
    publisher = "Association for Computational Linguistics",
    url = "https://aclanthology.org/2025.emnlp-main.1807/",
    doi = "10.18653/v1/2025.emnlp-main.1807",
    pages = "35673--35694",
    ISBN = "979-8-89176-332-6"
}

@inproceedings{zhang2025atomiccalibrationllmslongform,
    title = "Atomic Calibration of {LLM}s in Long-Form Generations",
    author = "Zhang, Caiqi  and
      Yang, Ruihan  and
      Zhang, Zhisong  and
      Huang, Xinting  and
      Yang, Sen  and
      Yu, Dong  and
      Collier, Nigel",
    editor = "Inui, Kentaro  and
      Sakti, Sakriani  and
      Wang, Haofen  and
      Wong, Derek F.  and
      Bhattacharyya, Pushpak  and
      Banerjee, Biplab  and
      Ekbal, Asif  and
      Chakraborty, Tanmoy  and
      Singh, Dhirendra Pratap",
    booktitle = "Proceedings of the 14th International Joint Conference on Natural Language Processing and the 4th Conference of the Asia-Pacific Chapter of the Association for Computational Linguistics",
    month = dec,
    year = "2025",
    address = "Mumbai, India",
    publisher = "The Asian Federation of Natural Language Processing and The Association for Computational Linguistics",
    url = "https://aclanthology.org/2025.findings-ijcnlp.9/",
    doi = "10.18653/v1/2025.findings-ijcnlp.9",
    pages = "148--169",
    ISBN = "979-8-89176-303-6"
}

@misc{bouchard2026functionalentropy,
      title={Functional Entropy: Predicting Functional Correctness in LLM-Generated Code with Uncertainty Quantification}, 
      author={Dylan Bouchard and Mohit Singh Chauhan and Zeya Ahmad and Ho-Kyeong Ra},
      year={2026},
      eprint={2605.28500},
      archivePrefix={arXiv},
      primaryClass={cs.CL},
      url={https://arxiv.org/abs/2605.28500}, 
}

@article{JMLR:v12:pedregosa11a,
  author  = {Fabian Pedregosa and Ga{{\"e}}l Varoquaux and Alexandre Gramfort and Vincent Michel and Bertrand Thirion and Olivier Grisel and Mathieu Blondel and Peter Prettenhofer and Ron Weiss and Vincent Dubourg and Jake Vanderplas and Alexandre Passos and David Cournapeau and Matthieu Brucher and Matthieu Perrot and {{\'E}}douard Duchesnay},
  title   = {Scikit-learn: Machine Learning in Python},
  journal = {Journal of Machine Learning Research},
  year    = {2011},
  volume  = {12},
  number  = {85},
  pages   = {2825--2830},
  url     = {http://jmlr.org/papers/v12/pedregosa11a.html}
}

@inproceedings{spiess2024calibrationcorrectnesslanguagemodels,
author = {Spiess, Claudio and Gros, David and Pai, Kunal Suresh and Pradel, Michael and Rabin, Md Rafiqul Islam and Alipour, Amin and Jha, Susmit and Devanbu, Prem and Ahmed, Toufique},
title = {Calibration and Correctness of Language Models for Code},
year = {2025},
isbn = {9798331505691},
publisher = {IEEE Press},
url = {https://doi.org/10.1109/ICSE55347.2025.00040},
doi = {10.1109/ICSE55347.2025.00040},
booktitle = {Proceedings of the IEEE/ACM 47th International Conference on Software Engineering},
pages = {540–552},
numpages = {13},
location = {Ottawa, Ontario, Canada},
series = {ICSE '25}
}

@inproceedings{vashurin2025uncertaintylinelengthinvariantestimationuncertainty,
    title = "{UNCERTAINTY}-{LINE}: Length-Invariant Estimation of Uncertainty for Large Language Models",
    author = "Vashurin, Roman  and
      Goloburda, Maiya  and
      Nakov, Preslav  and
      Panov, Maxim",
    editor = "Christodoulopoulos, Christos  and
      Chakraborty, Tanmoy  and
      Rose, Carolyn  and
      Peng, Violet",
    booktitle = "Proceedings of the 2025 Conference on Empirical Methods in Natural Language Processing",
    month = nov,
    year = "2025",
    address = "Suzhou, China",
    publisher = "Association for Computational Linguistics",
    url = "https://aclanthology.org/2025.emnlp-main.400/",
    doi = "10.18653/v1/2025.emnlp-main.400",
    pages = "7881--7908",
    ISBN = "979-8-89176-332-6"
}

@inproceedings{bakman2025reconsideringllmuncertaintyestimation,
    title = "Reconsidering {LLM} Uncertainty Estimation Methods in the Wild",
    author = "Bakman, Yavuz Faruk  and
      Yaldiz, Duygu Nur  and
      Kang, Sungmin  and
      Zhang, Tuo  and
      Buyukates, Baturalp  and
      Avestimehr, Salman  and
      Karimireddy, Sai Praneeth",
    editor = "Che, Wanxiang  and
      Nabende, Joyce  and
      Shutova, Ekaterina  and
      Pilehvar, Mohammad Taher",
    booktitle = "Proceedings of the 63rd Annual Meeting of the Association for Computational Linguistics (Volume 1: Long Papers)",
    month = jul,
    year = "2025",
    address = "Vienna, Austria",
    publisher = "Association for Computational Linguistics",
    url = "https://aclanthology.org/2025.acl-long.1429/",
    doi = "10.18653/v1/2025.acl-long.1429",
    pages = "29531--29556",
    ISBN = "979-8-89176-251-0"
}

@misc{sharma2025assessingcorrectnessllmbasedcode,
      title={Assessing Correctness in LLM-Based Code Generation via Uncertainty Estimation}, 
      author={Arindam Sharma and Cristina David},
      year={2025},
      eprint={2502.11620},
      archivePrefix={arXiv},
      primaryClass={cs.SE},
      url={https://arxiv.org/abs/2502.11620}, 
}

@misc{ren2020codebleumethodautomaticevaluation,
      title={CodeBLEU: a Method for Automatic Evaluation of Code Synthesis}, 
      author={Shuo Ren and Daya Guo and Shuai Lu and Long Zhou and Shujie Liu and Duyu Tang and Neel Sundaresan and Ming Zhou and Ambrosio Blanco and Shuai Ma},
      year={2020},
      eprint={2009.10297},
      archivePrefix={arXiv},
      primaryClass={cs.SE},
      url={https://arxiv.org/abs/2009.10297}, 
}

@misc{tian2023justaskcalibrationstrategies,
      title={Just Ask for Calibration: Strategies for Eliciting Calibrated Confidence Scores from Language Models Fine-Tuned with Human Feedback}, 
      author={Katherine Tian and Eric Mitchell and Allan Zhou and Archit Sharma and Rafael Rafailov and Huaxiu Yao and Chelsea Finn and Christopher D. Manning},
      year={2023},
      eprint={2305.14975},
      archivePrefix={arXiv},
      primaryClass={cs.CL},
      url={https://arxiv.org/abs/2305.14975}, 
}

@inproceedings{dua2019dropreadingcomprehensionbenchmark,
    title = "{DROP}: A Reading Comprehension Benchmark Requiring Discrete Reasoning Over Paragraphs",
    author = "Dua, Dheeru  and
      Wang, Yizhong  and
      Dasigi, Pradeep  and
      Stanovsky, Gabriel  and
      Singh, Sameer  and
      Gardner, Matt",
    editor = "Burstein, Jill  and
      Doran, Christy  and
      Solorio, Thamar",
    booktitle = "Proceedings of the 2019 Conference of the North {A}merican Chapter of the Association for Computational Linguistics: Human Language Technologies, Volume 1 (Long and Short Papers)",
    month = jun,
    year = "2019",
    address = "Minneapolis, Minnesota",
    publisher = "Association for Computational Linguistics",
    url = "https://aclanthology.org/N19-1246/",
    doi = "10.18653/v1/N19-1246",
    pages = "2368--2378"
}

@inproceedings{min2023factscorefinegrainedatomicevaluation,
    title = "{FA}ct{S}core: Fine-grained Atomic Evaluation of Factual Precision in Long Form Text Generation",
    author = "Min, Sewon  and
      Krishna, Kalpesh  and
      Lyu, Xinxi  and
      Lewis, Mike  and
      Yih, Wen-tau  and
      Koh, Pang  and
      Iyyer, Mohit  and
      Zettlemoyer, Luke  and
      Hajishirzi, Hannaneh",
    editor = "Bouamor, Houda  and
      Pino, Juan  and
      Bali, Kalika",
    booktitle = "Proceedings of the 2023 Conference on Empirical Methods in Natural Language Processing",
    month = dec,
    year = "2023",
    address = "Singapore",
    publisher = "Association for Computational Linguistics",
    url = "https://aclanthology.org/2023.emnlp-main.741/",
    doi = "10.18653/v1/2023.emnlp-main.741",
    pages = "12076--12100"
}

@inproceedings{yang2018hotpotqadatasetdiverseexplainable,
    title = "{H}otpot{QA}: A Dataset for Diverse, Explainable Multi-hop Question Answering",
    author = "Yang, Zhilin  and
      Qi, Peng  and
      Zhang, Saizheng  and
      Bengio, Yoshua  and
      Cohen, William  and
      Salakhutdinov, Ruslan  and
      Manning, Christopher D.",
    editor = "Riloff, Ellen  and
      Chiang, David  and
      Hockenmaier, Julia  and
      Tsujii, Jun{'}ichi",
    booktitle = "Proceedings of the 2018 Conference on Empirical Methods in Natural Language Processing",
    month = oct # "-" # nov,
    year = "2018",
    address = "Brussels, Belgium",
    publisher = "Association for Computational Linguistics",
    url = "https://aclanthology.org/D18-1259/",
    doi = "10.18653/v1/D18-1259",
    pages = "2369--2380"
}

@software{huggingface_open_r1_0_1_0_dev0,
  title        = {Open R1},
  author       = {{The Hugging Face team (past and future)}},
  year         = {2026},
  version      = {0.1.0.dev0},
  license      = {Apache-2.0},
  url          = {https://github.com/huggingface/open-r1},
  note         = {GitHub repository},
}

@inproceedings{
jain2024livecodebenchholisticcontaminationfree,
title={LiveCodeBench: Holistic and Contamination Free Evaluation of Large Language Models for Code},
author={Naman Jain and King Han and Alex Gu and Wen-Ding Li and Fanjia Yan and Tianjun Zhang and Sida Wang and Armando Solar-Lezama and Koushik Sen and Ion Stoica},
booktitle={The Thirteenth International Conference on Learning Representations},
year={2025},
url={https://openreview.net/forum?id=chfJJYC3iL}
}

@article{
bouchard2025uncertainty,
title={Uncertainty Quantification for Language Models: A Suite of Black-Box, White-Box, {LLM} Judge, and Ensemble Scorers},
author={Dylan Bouchard and Mohit Singh Chauhan},
journal={Transactions on Machine Learning Research},
issn={2835-8856},
year={2025},
url={https://openreview.net/forum?id=WOFspd4lq5},
note={}
}

@article{
bouchard2026finegraineduncertaintyquantificationlongform,
title={Fine-Grained Uncertainty Quantification for Long-Form Language Model Outputs: A Comparative Study},
author={Dylan Bouchard and Mohit Singh Chauhan and Viren Bajaj and David Skarbrevik},
journal={Transactions on Machine Learning Research},
issn={2835-8856},
year={2026},
url={https://openreview.net/forum?id=gngp4Zz9Sj},
note={}
}

@misc{scalena2025eagerentropyawaregenerationadaptive,
      title={EAGER: Entropy-Aware GEneRation for Adaptive Inference-Time Scaling}, 
      author={Daniel Scalena and Leonidas Zotos and Elisabetta Fersini and Malvina Nissim and Ahmet Üstün},
      year={2025},
      eprint={2510.11170},
      archivePrefix={arXiv},
      primaryClass={cs.LG},
      url={https://arxiv.org/abs/2510.11170}, 
}

@article{Vashurin_2025,
   title={Benchmarking Uncertainty Quantification Methods for Large Language Models with LM-Polygraph},
   volume={13},
   ISSN={2307-387X},
   url={http://dx.doi.org/10.1162/tacl_a_00737},
   DOI={10.1162/tacl_a_00737},
   journal={Transactions of the Association for Computational Linguistics},
   publisher={MIT Press},
   author={Vashurin, Roman and Fadeeva, Ekaterina and Vazhentsev, Artem and Rvanova, Lyudmila and Vasilev, Daniil and Tsvigun, Akim and Petrakov, Sergey and Xing, Rui and Sadallah, Abdelrahman and Grishchenkov, Kirill and Panchenko, Alexander and Baldwin, Timothy and Nakov, Preslav and Panov, Maxim and Shelmanov, Artem},
   year={2025},
   pages={220–248} }

@inproceedings{farr2024redctsystemsdesignmethodology,
    title = "{RED}-{CT}: A Systems Design Methodology for Using {LLM}-labeled Data to Train and Deploy Edge Linguistic Classifiers",
    author = "Farr, David  and
      Manzonelli, Nico  and
      Cruickshank, Iain  and
      West, Jevin",
    editor = "Rambow, Owen  and
      Wanner, Leo  and
      Apidianaki, Marianna  and
      Al-Khalifa, Hend  and
      Eugenio, Barbara Di  and
      Schockaert, Steven  and
      Darwish, Kareem  and
      Agarwal, Apoorv",
    booktitle = "Proceedings of the 31st International Conference on Computational Linguistics: Industry Track",
    month = jan,
    year = "2025",
    address = "Abu Dhabi, UAE",
    publisher = "Association for Computational Linguistics",
    url = "https://aclanthology.org/2025.coling-industry.5/",
    pages = "58--67"
}

@inproceedings{
vashurin2025uncertaintyquantificationllmsminimum,
title={CoCoA: A Minimum Bayes Risk Framework Bridging Confidence and Consistency for Uncertainty Quantification in {LLM}s},
author={Roman Vashurin and Maiya Goloburda and Albina Ilina and Aleksandr Rubashevskii and Preslav Nakov and Artem Shelmanov and Maxim Panov},
booktitle={The Thirty-ninth Annual Conference on Neural Information Processing Systems},
year={2025},
url={https://openreview.net/forum?id=H1NGlLNaVC}
}

@inproceedings{Akiba_Optuna_A_next-generation_2019,
author = {Akiba, Takuya and Sano, Shotaro and Yanase, Toshihiko and Ohta, Takeru and Koyama, Masanori},
booktitle = {Proceedings of the 25th ACM SIGKDD international conference on knowledge discovery \& data mining},
doi = {10.1145/3292500.3330701},
pages = {2623--2631},
title = {{Optuna: A next-generation hyperparameter optimization framework}},
year = {2019}
}

@misc{openai_doc,
      title={GPT-4o System Card}, 
      author={OpenAI and : and Aaron Hurst and Adam Lerer and Adam P. Goucher and Adam Perelman and Aditya Ramesh and Aidan Clark and AJ Ostrow and Akila Welihinda and Alan Hayes and Alec Radford and Aleksander Mądry and Alex Baker-Whitcomb and Alex Beutel and Alex Borzunov and Alex Carney and Alex Chow and Alex Kirillov and Alex Nichol and Alex Paino and Alex Renzin and Alex Tachard Passos and Alexander Kirillov and Alexi Christakis and Alexis Conneau and Ali Kamali and Allan Jabri and Allison Moyer and Allison Tam and Amadou Crookes and Amin Tootoochian and Amin Tootoonchian and Ananya Kumar and Andrea Vallone and Andrej Karpathy and Andrew Braunstein and Andrew Cann and Andrew Codispoti and Andrew Galu and Andrew Kondrich and Andrew Tulloch and Andrey Mishchenko and Angela Baek and Angela Jiang and Antoine Pelisse and Antonia Woodford and Anuj Gosalia and Arka Dhar and Ashley Pantuliano and Avi Nayak and Avital Oliver and Barret Zoph and Behrooz Ghorbani and Ben Leimberger and Ben Rossen and Ben Sokolowsky and Ben Wang and Benjamin Zweig and Beth Hoover and Blake Samic and Bob McGrew and Bobby Spero and Bogo Giertler and Bowen Cheng and Brad Lightcap and Brandon Walkin and Brendan Quinn and Brian Guarraci and Brian Hsu and Bright Kellogg and Brydon Eastman and Camillo Lugaresi and Carroll Wainwright and Cary Bassin and Cary Hudson and Casey Chu and Chad Nelson and Chak Li and Chan Jun Shern and Channing Conger and Charlotte Barette and Chelsea Voss and Chen Ding and Cheng Lu and Chong Zhang and Chris Beaumont and Chris Hallacy and Chris Koch and Christian Gibson and Christina Kim and Christine Choi and Christine McLeavey and Christopher Hesse and Claudia Fischer and Clemens Winter and Coley Czarnecki and Colin Jarvis and Colin Wei and Constantin Koumouzelis and Dane Sherburn and Daniel Kappler and Daniel Levin and Daniel Levy and David Carr and David Farhi and David Mely and David Robinson and David Sasaki and Denny Jin and Dev Valladares and Dimitris Tsipras and Doug Li and Duc Phong Nguyen and Duncan Findlay and Edede Oiwoh and Edmund Wong and Ehsan Asdar and Elizabeth Proehl and Elizabeth Yang and Eric Antonow and Eric Kramer and Eric Peterson and Eric Sigler and Eric Wallace and Eugene Brevdo and Evan Mays and Farzad Khorasani and Felipe Petroski Such and Filippo Raso and Francis Zhang and Fred von Lohmann and Freddie Sulit and Gabriel Goh and Gene Oden and Geoff Salmon and Giulio Starace and Greg Brockman and Hadi Salman and Haiming Bao and Haitang Hu and Hannah Wong and Haoyu Wang and Heather Schmidt and Heather Whitney and Heewoo Jun and Hendrik Kirchner and Henrique Ponde de Oliveira Pinto and Hongyu Ren and Huiwen Chang and Hyung Won Chung and Ian Kivlichan and Ian O'Connell and Ian O'Connell and Ian Osband and Ian Silber and Ian Sohl and Ibrahim Okuyucu and Ikai Lan and Ilya Kostrikov and Ilya Sutskever and Ingmar Kanitscheider and Ishaan Gulrajani and Jacob Coxon and Jacob Menick and Jakub Pachocki and James Aung and James Betker and James Crooks and James Lennon and Jamie Kiros and Jan Leike and Jane Park and Jason Kwon and Jason Phang and Jason Teplitz and Jason Wei and Jason Wolfe and Jay Chen and Jeff Harris and Jenia Varavva and Jessica Gan Lee and Jessica Shieh and Ji Lin and Jiahui Yu and Jiayi Weng and Jie Tang and Jieqi Yu and Joanne Jang and Joaquin Quinonero Candela and Joe Beutler and Joe Landers and Joel Parish and Johannes Heidecke and John Schulman and Jonathan Lachman and Jonathan McKay and Jonathan Uesato and Jonathan Ward and Jong Wook Kim and Joost Huizinga and Jordan Sitkin and Jos Kraaijeveld and Josh Gross and Josh Kaplan and Josh Snyder and Joshua Achiam and Joy Jiao and Joyce Lee and Juntang Zhuang and Justyn Harriman and Kai Fricke and Kai Hayashi and Karan Singhal and Katy Shi and Kavin Karthik and Kayla Wood and Kendra Rimbach and Kenny Hsu and Kenny Nguyen and Keren Gu-Lemberg and Kevin Button and Kevin Liu and Kiel Howe and Krithika Muthukumar and Kyle Luther and Lama Ahmad and Larry Kai and Lauren Itow and Lauren Workman and Leher Pathak and Leo Chen and Li Jing and Lia Guy and Liam Fedus and Liang Zhou and Lien Mamitsuka and Lilian Weng and Lindsay McCallum and Lindsey Held and Long Ouyang and Louis Feuvrier and Lu Zhang and Lukas Kondraciuk and Lukasz Kaiser and Luke Hewitt and Luke Metz and Lyric Doshi and Mada Aflak and Maddie Simens and Madelaine Boyd and Madeleine Thompson and Marat Dukhan and Mark Chen and Mark Gray and Mark Hudnall and Marvin Zhang and Marwan Aljubeh and Mateusz Litwin and Matthew Zeng and Max Johnson and Maya Shetty and Mayank Gupta and Meghan Shah and Mehmet Yatbaz and Meng Jia Yang and Mengchao Zhong and Mia Glaese and Mianna Chen and Michael Janner and Michael Lampe and Michael Petrov and Michael Wu and Michele Wang and Michelle Fradin and Michelle Pokrass and Miguel Castro and Miguel Oom Temudo de Castro and Mikhail Pavlov and Miles Brundage and Miles Wang and Minal Khan and Mira Murati and Mo Bavarian and Molly Lin and Murat Yesildal and Nacho Soto and Natalia Gimelshein and Natalie Cone and Natalie Staudacher and Natalie Summers and Natan LaFontaine and Neil Chowdhury and Nick Ryder and Nick Stathas and Nick Turley and Nik Tezak and Niko Felix and Nithanth Kudige and Nitish Keskar and Noah Deutsch and Noel Bundick and Nora Puckett and Ofir Nachum and Ola Okelola and Oleg Boiko and Oleg Murk and Oliver Jaffe and Olivia Watkins and Olivier Godement and Owen Campbell-Moore and Patrick Chao and Paul McMillan and Pavel Belov and Peng Su and Peter Bak and Peter Bakkum and Peter Deng and Peter Dolan and Peter Hoeschele and Peter Welinder and Phil Tillet and Philip Pronin and Philippe Tillet and Prafulla Dhariwal and Qiming Yuan and Rachel Dias and Rachel Lim and Rahul Arora and Rajan Troll and Randall Lin and Rapha Gontijo Lopes and Raul Puri and Reah Miyara and Reimar Leike and Renaud Gaubert and Reza Zamani and Ricky Wang and Rob Donnelly and Rob Honsby and Rocky Smith and Rohan Sahai and Rohit Ramchandani and Romain Huet and Rory Carmichael and Rowan Zellers and Roy Chen and Ruby Chen and Ruslan Nigmatullin and Ryan Cheu and Saachi Jain and Sam Altman and Sam Schoenholz and Sam Toizer and Samuel Miserendino and Sandhini Agarwal and Sara Culver and Scott Ethersmith and Scott Gray and Sean Grove and Sean Metzger and Shamez Hermani and Shantanu Jain and Shengjia Zhao and Sherwin Wu and Shino Jomoto and Shirong Wu and Shuaiqi and Xia and Sonia Phene and Spencer Papay and Srinivas Narayanan and Steve Coffey and Steve Lee and Stewart Hall and Suchir Balaji and Tal Broda and Tal Stramer and Tao Xu and Tarun Gogineni and Taya Christianson and Ted Sanders and Tejal Patwardhan and Thomas Cunninghman and Thomas Degry and Thomas Dimson and Thomas Raoux and Thomas Shadwell and Tianhao Zheng and Todd Underwood and Todor Markov and Toki Sherbakov and Tom Rubin and Tom Stasi and Tomer Kaftan and Tristan Heywood and Troy Peterson and Tyce Walters and Tyna Eloundou and Valerie Qi and Veit Moeller and Vinnie Monaco and Vishal Kuo and Vlad Fomenko and Wayne Chang and Weiyi Zheng and Wenda Zhou and Wesam Manassra and Will Sheu and Wojciech Zaremba and Yash Patil and Yilei Qian and Yongjik Kim and Youlong Cheng and Yu Zhang and Yuchen He and Yuchen Zhang and Yujia Jin and Yunxing Dai and Yury Malkov},
      year={2024},
      eprint={2410.21276},
      archivePrefix={arXiv},
      primaryClass={cs.CL},
      url={https://arxiv.org/abs/2410.21276}, 
}

@misc{gemini_doc,
      title={Gemini: A Family of Highly Capable Multimodal Models}, 
      author={{Gemini Team} and Rohan Anil and Sebastian Borgeaud and Jean-Baptiste Alayrac and Jiahui Yu and Radu Soricut and Johan Schalkwyk and Andrew M. Dai and Anja Hauth and Katie Millican and David Silver and Melvin Johnson and Ioannis Antonoglou and Julian Schrittwieser and Amelia Glaese and Jilin Chen and Emily Pitler and Timothy Lillicrap and Angeliki Lazaridou and Orhan Firat and James Molloy and Michael Isard and Paul R. Barham and Tom Hennigan and Benjamin Lee and Fabio Viola and Malcolm Reynolds and Yuanzhong Xu and Ryan Doherty and Eli Collins and Clemens Meyer and Eliza Rutherford and Erica Moreira and Kareem Ayoub and Megha Goel and Jack Krawczyk and Cosmo Du and Ed Chi and Heng-Tze Cheng and Eric Ni and Purvi Shah and Patrick Kane and Betty Chan and Manaal Faruqui and Aliaksei Severyn and Hanzhao Lin and YaGuang Li and Yong Cheng and Abe Ittycheriah and Mahdis Mahdieh and Mia Chen and Pei Sun and Dustin Tran and Sumit Bagri and Balaji Lakshminarayanan and Jeremiah Liu and Andras Orban and Fabian Güra and Hao Zhou and Xinying Song and Aurelien Boffy and Harish Ganapathy and Steven Zheng and HyunJeong Choe and Ágoston Weisz and Tao Zhu and Yifeng Lu and Siddharth Gopal and Jarrod Kahn and Maciej Kula and Jeff Pitman and Rushin Shah and Emanuel Taropa and Majd Al Merey and Martin Baeuml and Zhifeng Chen and Laurent El Shafey and Yujing Zhang and Olcan Sercinoglu and George Tucker and Enrique Piqueras and Maxim Krikun and Iain Barr and Nikolay Savinov and Ivo Danihelka and Becca Roelofs and Anaïs White and Anders Andreassen and Tamara von Glehn and Lakshman Yagati and Mehran Kazemi and Lucas Gonzalez and Misha Khalman and Jakub Sygnowski and Alexandre Frechette and Charlotte Smith and Laura Culp and Lev Proleev and Yi Luan and Xi Chen and James Lottes and Nathan Schucher and Federico Lebron and Alban Rrustemi and Natalie Clay and Phil Crone and Tomas Kocisky and Jeffrey Zhao and Bartek Perz and Dian Yu and Heidi Howard and Adam Bloniarz and Jack W. Rae and Han Lu and Laurent Sifre and Marcello Maggioni and Fred Alcober and Dan Garrette and Megan Barnes and Shantanu Thakoor and Jacob Austin and Gabriel Barth-Maron and William Wong and Rishabh Joshi and Rahma Chaabouni and Deeni Fatiha and Arun Ahuja and Gaurav Singh Tomar and Evan Senter and Martin Chadwick and Ilya Kornakov and Nithya Attaluri and Iñaki Iturrate and Ruibo Liu and Yunxuan Li and Sarah Cogan and Jeremy Chen and Chao Jia and Chenjie Gu and Qiao Zhang and Jordan Grimstad and Ale Jakse Hartman and Xavier Garcia and Thanumalayan Sankaranarayana Pillai and Jacob Devlin and Michael Laskin and Diego de Las Casas and Dasha Valter and Connie Tao and Lorenzo Blanco and Adrià Puigdomènech Badia and David Reitter and Mianna Chen and Jenny Brennan and Clara Rivera and Sergey Brin and Shariq Iqbal and Gabriela Surita and Jane Labanowski and Abhi Rao and Stephanie Winkler and Emilio Parisotto and Yiming Gu and Kate Olszewska and Ravi Addanki and Antoine Miech and Annie Louis and Denis Teplyashin and Geoff Brown and Elliot Catt and Jan Balaguer and Jackie Xiang and Pidong Wang and Zoe Ashwood and Anton Briukhov and Albert Webson and Sanjay Ganapathy and Smit Sanghavi and Ajay Kannan and Ming-Wei Chang and Axel Stjerngren and Josip Djolonga and Yuting Sun and Ankur Bapna and Matthew Aitchison and Pedram Pejman and Henryk Michalewski and Tianhe Yu and Cindy Wang and Juliette Love and Junwhan Ahn and Dawn Bloxwich and Kehang Han and Peter Humphreys and Thibault Sellam and James Bradbury and Varun Godbole and Sina Samangooei and Bogdan Damoc and Alex Kaskasoli and Sébastien M. R. Arnold and Vijay Vasudevan and Shubham Agrawal and Jason Riesa and Dmitry Lepikhin and Richard Tanburn and Srivatsan Srinivasan and Hyeontaek Lim and Sarah Hodkinson and Pranav Shyam and Johan Ferret and Steven Hand and Ankush Garg and Tom Le Paine and Jian Li and Yujia Li and Minh Giang and Alexander Neitz and Zaheer Abbas and Sarah York and Machel Reid and Elizabeth Cole and Aakanksha Chowdhery and Dipanjan Das and Dominika Rogozińska and Vitaliy Nikolaev and Pablo Sprechmann and Zachary Nado and Lukas Zilka and Flavien Prost and Luheng He and Marianne Monteiro and Gaurav Mishra and Chris Welty and Josh Newlan and Dawei Jia and Miltiadis Allamanis and Clara Huiyi Hu and Raoul de Liedekerke and Justin Gilmer and Carl Saroufim and Shruti Rijhwani and Shaobo Hou and Disha Shrivastava and Anirudh Baddepudi and Alex Goldin and Adnan Ozturel and Albin Cassirer and Yunhan Xu and Daniel Sohn and Devendra Sachan and Reinald Kim Amplayo and Craig Swanson and Dessie Petrova and Shashi Narayan and Arthur Guez and Siddhartha Brahma and Jessica Landon and Miteyan Patel and Ruizhe Zhao and Kevin Villela and Luyu Wang and Wenhao Jia and Matthew Rahtz and Mai Giménez and Legg Yeung and James Keeling and Petko Georgiev and Diana Mincu and Boxi Wu and Salem Haykal and Rachel Saputro and Kiran Vodrahalli and James Qin and Zeynep Cankara and Abhanshu Sharma and Nick Fernando and Will Hawkins and Behnam Neyshabur and Solomon Kim and Adrian Hutter and Priyanka Agrawal and Alex Castro-Ros and George van den Driessche and Tao Wang and Fan Yang and Shuo-yiin Chang and Paul Komarek and Ross McIlroy and Mario Lučić and Guodong Zhang and Wael Farhan and Michael Sharman and Paul Natsev and Paul Michel and Yamini Bansal and Siyuan Qiao and Kris Cao and Siamak Shakeri and Christina Butterfield and Justin Chung and Paul Kishan Rubenstein and Shivani Agrawal and Arthur Mensch and Kedar Soparkar and Karel Lenc and Timothy Chung and Aedan Pope and Loren Maggiore and Jackie Kay and Priya Jhakra and Shibo Wang and Joshua Maynez and Mary Phuong and Taylor Tobin and Andrea Tacchetti and Maja Trebacz and Kevin Robinson and Yash Katariya and Sebastian Riedel and Paige Bailey and Kefan Xiao and Nimesh Ghelani and Lora Aroyo and Ambrose Slone and Neil Houlsby and Xuehan Xiong and Zhen Yang and Elena Gribovskaya and Jonas Adler and Mateo Wirth and Lisa Lee and Music Li and Thais Kagohara and Jay Pavagadhi and Sophie Bridgers and Anna Bortsova and Sanjay Ghemawat and Zafarali Ahmed and Tianqi Liu and Richard Powell and Vijay Bolina and Mariko Iinuma and Polina Zablotskaia and James Besley and Da-Woon Chung and Timothy Dozat and Ramona Comanescu and Xiance Si and Jeremy Greer and Guolong Su and Martin Polacek and Raphaël Lopez Kaufman and Simon Tokumine and Hexiang Hu and Elena Buchatskaya and Yingjie Miao and Mohamed Elhawaty and Aditya Siddhant and Nenad Tomasev and Jinwei Xing and Christina Greer and Helen Miller and Shereen Ashraf and Aurko Roy and Zizhao Zhang and Ada Ma and Angelos Filos and Milos Besta and Rory Blevins and Ted Klimenko and Chih-Kuan Yeh and Soravit Changpinyo and Jiaqi Mu and Oscar Chang and Mantas Pajarskas and Carrie Muir and Vered Cohen and Charline Le Lan and Krishna Haridasan and Amit Marathe and Steven Hansen and Sholto Douglas and Rajkumar Samuel and Mingqiu Wang and Sophia Austin and Chang Lan and Jiepu Jiang and Justin Chiu and Jaime Alonso Lorenzo and Lars Lowe Sjösund and Sébastien Cevey and Zach Gleicher and Thi Avrahami and Anudhyan Boral and Hansa Srinivasan and Vittorio Selo and Rhys May and Konstantinos Aisopos and Léonard Hussenot and Livio Baldini Soares and Kate Baumli and Michael B. Chang and Adrià Recasens and Ben Caine and Alexander Pritzel and Filip Pavetic and Fabio Pardo and Anita Gergely and Justin Frye and Vinay Ramasesh and Dan Horgan and Kartikeya Badola and Nora Kassner and Subhrajit Roy and Ethan Dyer and Víctor Campos Campos and Alex Tomala and Yunhao Tang and Dalia El Badawy and Elspeth White and Basil Mustafa and Oran Lang and Abhishek Jindal and Sharad Vikram and Zhitao Gong and Sergi Caelles and Ross Hemsley and Gregory Thornton and Fangxiaoyu Feng and Wojciech Stokowiec and Ce Zheng and Phoebe Thacker and Çağlar Ünlü and Zhishuai Zhang and Mohammad Saleh and James Svensson and Max Bileschi and Piyush Patil and Ankesh Anand and Roman Ring and Katerina Tsihlas and Arpi Vezer and Marco Selvi and Toby Shevlane and Mikel Rodriguez and Tom Kwiatkowski and Samira Daruki and Keran Rong and Allan Dafoe and Nicholas FitzGerald and Keren Gu-Lemberg and Mina Khan and Lisa Anne Hendricks and Marie Pellat and Vladimir Feinberg and James Cobon-Kerr and Tara Sainath and Maribeth Rauh and Sayed Hadi Hashemi and Richard Ives and Yana Hasson and Eric Noland and Yuan Cao and Nathan Byrd and Le Hou and Qingze Wang and Thibault Sottiaux and Michela Paganini and Jean-Baptiste Lespiau and Alexandre Moufarek and Samer Hassan and Kaushik Shivakumar and Joost van Amersfoort and Amol Mandhane and Pratik Joshi and Anirudh Goyal and Matthew Tung and Andrew Brock and Hannah Sheahan and Vedant Misra and Cheng Li and Nemanja Rakićević and Mostafa Dehghani and Fangyu Liu and Sid Mittal and Junhyuk Oh and Seb Noury and Eren Sezener and Fantine Huot and Matthew Lamm and Nicola De Cao and Charlie Chen and Sidharth Mudgal and Romina Stella and Kevin Brooks and Gautam Vasudevan and Chenxi Liu and Mainak Chain and Nivedita Melinkeri and Aaron Cohen and Venus Wang and Kristie Seymore and Sergey Zubkov and Rahul Goel and Summer Yue and Sai Krishnakumaran and Brian Albert and Nate Hurley and Motoki Sano and Anhad Mohananey and Jonah Joughin and Egor Filonov and Tomasz Kępa and Yomna Eldawy and Jiawern Lim and Rahul Rishi and Shirin Badiezadegan and Taylor Bos and Jerry Chang and Sanil Jain and Sri Gayatri Sundara Padmanabhan and Subha Puttagunta and Kalpesh Krishna and Leslie Baker and Norbert Kalb and Vamsi Bedapudi and Adam Kurzrok and Shuntong Lei and Anthony Yu and Oren Litvin and Xiang Zhou and Zhichun Wu and Sam Sobell and Andrea Siciliano and Alan Papir and Robby Neale and Jonas Bragagnolo and Tej Toor and Tina Chen and Valentin Anklin and Feiran Wang and Richie Feng and Milad Gholami and Kevin Ling and Lijuan Liu and Jules Walter and Hamid Moghaddam and Arun Kishore and Jakub Adamek and Tyler Mercado and Jonathan Mallinson and Siddhinita Wandekar and Stephen Cagle and Eran Ofek and Guillermo Garrido and Clemens Lombriser and Maksim Mukha and Botu Sun and Hafeezul Rahman Mohammad and Josip Matak and Yadi Qian and Vikas Peswani and Pawel Janus and Quan Yuan and Leif Schelin and Oana David and Ankur Garg and Yifan He and Oleksii Duzhyi and Anton Älgmyr and Timothée Lottaz and Qi Li and Vikas Yadav and Luyao Xu and Alex Chinien and Rakesh Shivanna and Aleksandr Chuklin and Josie Li and Carrie Spadine and Travis Wolfe and Kareem Mohamed and Subhabrata Das and Zihang Dai and Kyle He and Daniel von Dincklage and Shyam Upadhyay and Akanksha Maurya and Luyan Chi and Sebastian Krause and Khalid Salama and Pam G Rabinovitch and Pavan Kumar Reddy M and Aarush Selvan and Mikhail Dektiarev and Golnaz Ghiasi and Erdem Guven and Himanshu Gupta and Boyi Liu and Deepak Sharma and Idan Heimlich Shtacher and Shachi Paul and Oscar Akerlund and François-Xavier Aubet and Terry Huang and Chen Zhu and Eric Zhu and Elico Teixeira and Matthew Fritze and Francesco Bertolini and Liana-Eleonora Marinescu and Martin Bölle and Dominik Paulus and Khyatti Gupta and Tejasi Latkar and Max Chang and Jason Sanders and Roopa Wilson and Xuewei Wu and Yi-Xuan Tan and Lam Nguyen Thiet and Tulsee Doshi and Sid Lall and Swaroop Mishra and Wanming Chen and Thang Luong and Seth Benjamin and Jasmine Lee and Ewa Andrejczuk and Dominik Rabiej and Vipul Ranjan and Krzysztof Styrc and Pengcheng Yin and Jon Simon and Malcolm Rose Harriott and Mudit Bansal and Alexei Robsky and Geoff Bacon and David Greene and Daniil Mirylenka and Chen Zhou and Obaid Sarvana and Abhimanyu Goyal and Samuel Andermatt and Patrick Siegler and Ben Horn and Assaf Israel and Francesco Pongetti and Chih-Wei "Louis" Chen and Marco Selvatici and Pedro Silva and Kathie Wang and Jackson Tolins and Kelvin Guu and Roey Yogev and Xiaochen Cai and Alessandro Agostini and Maulik Shah and Hung Nguyen and Noah Ó Donnaile and Sébastien Pereira and Linda Friso and Adam Stambler and Adam Kurzrok and Chenkai Kuang and Yan Romanikhin and Mark Geller and ZJ Yan and Kane Jang and Cheng-Chun Lee and Wojciech Fica and Eric Malmi and Qijun Tan and Dan Banica and Daniel Balle and Ryan Pham and Yanping Huang and Diana Avram and Hongzhi Shi and Jasjot Singh and Chris Hidey and Niharika Ahuja and Pranab Saxena and Dan Dooley and Srividya Pranavi Potharaju and Eileen O'Neill and Anand Gokulchandran and Ryan Foley and Kai Zhao and Mike Dusenberry and Yuan Liu and Pulkit Mehta and Ragha Kotikalapudi and Chalence Safranek-Shrader and Andrew Goodman and Joshua Kessinger and Eran Globen and Prateek Kolhar and Chris Gorgolewski and Ali Ibrahim and Yang Song and Ali Eichenbaum and Thomas Brovelli and Sahitya Potluri and Preethi Lahoti and Cip Baetu and Ali Ghorbani and Charles Chen and Andy Crawford and Shalini Pal and Mukund Sridhar and Petru Gurita and Asier Mujika and Igor Petrovski and Pierre-Louis Cedoz and Chenmei Li and Shiyuan Chen and Niccolò Dal Santo and Siddharth Goyal and Jitesh Punjabi and Karthik Kappaganthu and Chester Kwak and Pallavi LV and Sarmishta Velury and Himadri Choudhury and Jamie Hall and Premal Shah and Ricardo Figueira and Matt Thomas and Minjie Lu and Ting Zhou and Chintu Kumar and Thomas Jurdi and Sharat Chikkerur and Yenai Ma and Adams Yu and Soo Kwak and Victor Ähdel and Sujeevan Rajayogam and Travis Choma and Fei Liu and Aditya Barua and Colin Ji and Ji Ho Park and Vincent Hellendoorn and Alex Bailey and Taylan Bilal and Huanjie Zhou and Mehrdad Khatir and Charles Sutton and Wojciech Rzadkowski and Fiona Macintosh and Roopali Vij and Konstantin Shagin and Paul Medina and Chen Liang and Jinjing Zhou and Pararth Shah and Yingying Bi and Attila Dankovics and Shipra Banga and Sabine Lehmann and Marissa Bredesen and Zifan Lin and John Eric Hoffmann and Jonathan Lai and Raynald Chung and Kai Yang and Nihal Balani and Arthur Bražinskas and Andrei Sozanschi and Matthew Hayes and Héctor Fernández Alcalde and Peter Makarov and Will Chen and Antonio Stella and Liselotte Snijders and Michael Mandl and Ante Kärrman and Paweł Nowak and Xinyi Wu and Alex Dyck and Krishnan Vaidyanathan and Raghavender R and Jessica Mallet and Mitch Rudominer and Eric Johnston and Sushil Mittal and Akhil Udathu and Janara Christensen and Vishal Verma and Zach Irving and Andreas Santucci and Gamaleldin Elsayed and Elnaz Davoodi and Marin Georgiev and Ian Tenney and Nan Hua and Geoffrey Cideron and Edouard Leurent and Mahmoud Alnahlawi and Ionut Georgescu and Nan Wei and Ivy Zheng and Dylan Scandinaro and Heinrich Jiang and Jasper Snoek and Mukund Sundararajan and Xuezhi Wang and Zack Ontiveros and Itay Karo and Jeremy Cole and Vinu Rajashekhar and Lara Tumeh and Eyal Ben-David and Rishub Jain and Jonathan Uesato and Romina Datta and Oskar Bunyan and Shimu Wu and John Zhang and Piotr Stanczyk and Ye Zhang and David Steiner and Subhajit Naskar and Michael Azzam and Matthew Johnson and Adam Paszke and Chung-Cheng Chiu and Jaume Sanchez Elias and Afroz Mohiuddin and Faizan Muhammad and Jin Miao and Andrew Lee and Nino Vieillard and Jane Park and Jiageng Zhang and Jeff Stanway and Drew Garmon and Abhijit Karmarkar and Zhe Dong and Jong Lee and Aviral Kumar and Luowei Zhou and Jonathan Evens and William Isaac and Geoffrey Irving and Edward Loper and Michael Fink and Isha Arkatkar and Nanxin Chen and Izhak Shafran and Ivan Petrychenko and Zhe Chen and Johnson Jia and Anselm Levskaya and Zhenkai Zhu and Peter Grabowski and Yu Mao and Alberto Magni and Kaisheng Yao and Javier Snaider and Norman Casagrande and Evan Palmer and Paul Suganthan and Alfonso Castaño and Irene Giannoumis and Wooyeol Kim and Mikołaj Rybiński and Ashwin Sreevatsa and Jennifer Prendki and David Soergel and Adrian Goedeckemeyer and Willi Gierke and Mohsen Jafari and Meenu Gaba and Jeremy Wiesner and Diana Gage Wright and Yawen Wei and Harsha Vashisht and Yana Kulizhskaya and Jay Hoover and Maigo Le and Lu Li and Chimezie Iwuanyanwu and Lu Liu and Kevin Ramirez and Andrey Khorlin and Albert Cui and Tian LIN and Marcus Wu and Ricardo Aguilar and Keith Pallo and Abhishek Chakladar and Ginger Perng and Elena Allica Abellan and Mingyang Zhang and Ishita Dasgupta and Nate Kushman and Ivo Penchev and Alena Repina and Xihui Wu and Tom van der Weide and Priya Ponnapalli and Caroline Kaplan and Jiri Simsa and Shuangfeng Li and Olivier Dousse and Fan Yang and Jeff Piper and Nathan Ie and Rama Pasumarthi and Nathan Lintz and Anitha Vijayakumar and Daniel Andor and Pedro Valenzuela and Minnie Lui and Cosmin Paduraru and Daiyi Peng and Katherine Lee and Shuyuan Zhang and Somer Greene and Duc Dung Nguyen and Paula Kurylowicz and Cassidy Hardin and Lucas Dixon and Lili Janzer and Kiam Choo and Ziqiang Feng and Biao Zhang and Achintya Singhal and Dayou Du and Dan McKinnon and Natasha Antropova and Tolga Bolukbasi and Orgad Keller and David Reid and Daniel Finchelstein and Maria Abi Raad and Remi Crocker and Peter Hawkins and Robert Dadashi and Colin Gaffney and Ken Franko and Anna Bulanova and Rémi Leblond and Shirley Chung and Harry Askham and Luis C. Cobo and Kelvin Xu and Felix Fischer and Jun Xu and Christina Sorokin and Chris Alberti and Chu-Cheng Lin and Colin Evans and Alek Dimitriev and Hannah Forbes and Dylan Banarse and Zora Tung and Mark Omernick and Colton Bishop and Rachel Sterneck and Rohan Jain and Jiawei Xia and Ehsan Amid and Francesco Piccinno and Xingyu Wang and Praseem Banzal and Daniel J. Mankowitz and Alex Polozov and Victoria Krakovna and Sasha Brown and MohammadHossein Bateni and Dennis Duan and Vlad Firoiu and Meghana Thotakuri and Tom Natan and Matthieu Geist and Ser tan Girgin and Hui Li and Jiayu Ye and Ofir Roval and Reiko Tojo and Michael Kwong and James Lee-Thorp and Christopher Yew and Danila Sinopalnikov and Sabela Ramos and John Mellor and Abhishek Sharma and Kathy Wu and David Miller and Nicolas Sonnerat and Denis Vnukov and Rory Greig and Jennifer Beattie and Emily Caveness and Libin Bai and Julian Eisenschlos and Alex Korchemniy and Tomy Tsai and Mimi Jasarevic and Weize Kong and Phuong Dao and Zeyu Zheng and Frederick Liu and Fan Yang and Rui Zhu and Tian Huey Teh and Jason Sanmiya and Evgeny Gladchenko and Nejc Trdin and Daniel Toyama and Evan Rosen and Sasan Tavakkol and Linting Xue and Chen Elkind and Oliver Woodman and John Carpenter and George Papamakarios and Rupert Kemp and Sushant Kafle and Tanya Grunina and Rishika Sinha and Alice Talbert and Diane Wu and Denese Owusu-Afriyie and Cosmo Du and Chloe Thornton and Jordi Pont-Tuset and Pradyumna Narayana and Jing Li and Saaber Fatehi and John Wieting and Omar Ajmeri and Benigno Uria and Yeongil Ko and Laura Knight and Amélie Héliou and Ning Niu and Shane Gu and Chenxi Pang and Yeqing Li and Nir Levine and Ariel Stolovich and Rebeca Santamaria-Fernandez and Sonam Goenka and Wenny Yustalim and Robin Strudel and Ali Elqursh and Charlie Deck and Hyo Lee and Zonglin Li and Kyle Levin and Raphael Hoffmann and Dan Holtmann-Rice and Olivier Bachem and Sho Arora and Christy Koh and Soheil Hassas Yeganeh and Siim Põder and Mukarram Tariq and Yanhua Sun and Lucian Ionita and Mojtaba Seyedhosseini and Pouya Tafti and Zhiyu Liu and Anmol Gulati and Jasmine Liu and Xinyu Ye and Bart Chrzaszcz and Lily Wang and Nikhil Sethi and Tianrun Li and Ben Brown and Shreya Singh and Wei Fan and Aaron Parisi and Joe Stanton and Vinod Koverkathu and Christopher A. Choquette-Choo and Yunjie Li and TJ Lu and Abe Ittycheriah and Prakash Shroff and Mani Varadarajan and Sanaz Bahargam and Rob Willoughby and David Gaddy and Guillaume Desjardins and Marco Cornero and Brona Robenek and Bhavishya Mittal and Ben Albrecht and Ashish Shenoy and Fedor Moiseev and Henrik Jacobsson and Alireza Ghaffarkhah and Morgane Rivière and Alanna Walton and Clément Crepy and Alicia Parrish and Zongwei Zhou and Clement Farabet and Carey Radebaugh and Praveen Srinivasan and Claudia van der Salm and Andreas Fidjeland and Salvatore Scellato and Eri Latorre-Chimoto and Hanna Klimczak-Plucińska and David Bridson and Dario de Cesare and Tom Hudson and Piermaria Mendolicchio and Lexi Walker and Alex Morris and Matthew Mauger and Alexey Guseynov and Alison Reid and Seth Odoom and Lucia Loher and Victor Cotruta and Madhavi Yenugula and Dominik Grewe and Anastasia Petrushkina and Tom Duerig and Antonio Sanchez and Steve Yadlowsky and Amy Shen and Amir Globerson and Lynette Webb and Sahil Dua and Dong Li and Surya Bhupatiraju and Dan Hurt and Haroon Qureshi and Ananth Agarwal and Tomer Shani and Matan Eyal and Anuj Khare and Shreyas Rammohan Belle and Lei Wang and Chetan Tekur and Mihir Sanjay Kale and Jinliang Wei and Ruoxin Sang and Brennan Saeta and Tyler Liechty and Yi Sun and Yao Zhao and Stephan Lee and Pandu Nayak and Doug Fritz and Manish Reddy Vuyyuru and John Aslanides and Nidhi Vyas and Martin Wicke and Xiao Ma and Evgenii Eltyshev and Nina Martin and Hardie Cate and James Manyika and Keyvan Amiri and Yelin Kim and Xi Xiong and Kai Kang and Florian Luisier and Nilesh Tripuraneni and David Madras and Mandy Guo and Austin Waters and Oliver Wang and Joshua Ainslie and Jason Baldridge and Han Zhang and Garima Pruthi and Jakob Bauer and Feng Yang and Riham Mansour and Jason Gelman and Yang Xu and George Polovets and Ji Liu and Honglong Cai and Warren Chen and XiangHai Sheng and Emily Xue and Sherjil Ozair and Christof Angermueller and Xiaowei Li and Anoop Sinha and Weiren Wang and Julia Wiesinger and Emmanouil Koukoumidis and Yuan Tian and Anand Iyer and Madhu Gurumurthy and Mark Goldenson and Parashar Shah and MK Blake and Hongkun Yu and Anthony Urbanowicz and Jennimaria Palomaki and Chrisantha Fernando and Ken Durden and Harsh Mehta and Nikola Momchev and Elahe Rahimtoroghi and Maria Georgaki and Amit Raul and Sebastian Ruder and Morgan Redshaw and Jinhyuk Lee and Denny Zhou and Komal Jalan and Dinghua Li and Blake Hechtman and Parker Schuh and Milad Nasr and Kieran Milan and Vladimir Mikulik and Juliana Franco and Tim Green and Nam Nguyen and Joe Kelley and Aroma Mahendru and Andrea Hu and Joshua Howland and Ben Vargas and Jeffrey Hui and Kshitij Bansal and Vikram Rao and Rakesh Ghiya and Emma Wang and Ke Ye and Jean Michel Sarr and Melanie Moranski Preston and Madeleine Elish and Steve Li and Aakash Kaku and Jigar Gupta and Ice Pasupat and Da-Cheng Juan and Milan Someswar and Tejvi M. and Xinyun Chen and Aida Amini and Alex Fabrikant and Eric Chu and Xuanyi Dong and Amruta Muthal and Senaka Buthpitiya and Sarthak Jauhari and Nan Hua and Urvashi Khandelwal and Ayal Hitron and Jie Ren and Larissa Rinaldi and Shahar Drath and Avigail Dabush and Nan-Jiang Jiang and Harshal Godhia and Uli Sachs and Anthony Chen and Yicheng Fan and Hagai Taitelbaum and Hila Noga and Zhuyun Dai and James Wang and Chen Liang and Jenny Hamer and Chun-Sung Ferng and Chenel Elkind and Aviel Atias and Paulina Lee and Vít Listík and Mathias Carlen and Jan van de Kerkhof and Marcin Pikus and Krunoslav Zaher and Paul Müller and Sasha Zykova and Richard Stefanec and Vitaly Gatsko and Christoph Hirnschall and Ashwin Sethi and Xingyu Federico Xu and Chetan Ahuja and Beth Tsai and Anca Stefanoiu and Bo Feng and Keshav Dhandhania and Manish Katyal and Akshay Gupta and Atharva Parulekar and Divya Pitta and Jing Zhao and Vivaan Bhatia and Yashodha Bhavnani and Omar Alhadlaq and Xiaolin Li and Peter Danenberg and Dennis Tu and Alex Pine and Vera Filippova and Abhipso Ghosh and Ben Limonchik and Bhargava Urala and Chaitanya Krishna Lanka and Derik Clive and Yi Sun and Edward Li and Hao Wu and Kevin Hongtongsak and Ianna Li and Kalind Thakkar and Kuanysh Omarov and Kushal Majmundar and Michael Alverson and Michael Kucharski and Mohak Patel and Mudit Jain and Maksim Zabelin and Paolo Pelagatti and Rohan Kohli and Saurabh Kumar and Joseph Kim and Swetha Sankar and Vineet Shah and Lakshmi Ramachandruni and Xiangkai Zeng and Ben Bariach and Laura Weidinger and Tu Vu and Alek Andreev and Antoine He and Kevin Hui and Sheleem Kashem and Amar Subramanya and Sissie Hsiao and Demis Hassabis and Koray Kavukcuoglu and Adam Sadovsky and Quoc Le and Trevor Strohman and Yonghui Wu and Slav Petrov and Jeffrey Dean and Oriol Vinyals},
      year={2025},
      eprint={2312.11805},
      archivePrefix={arXiv},
      primaryClass={cs.CL},
      url={https://arxiv.org/abs/2312.11805}, 
}

@article{bouchard2025uqlmpythonpackageuncertainty,
  author  = {Dylan Bouchard and Mohit Singh Chauhan and David Skarbrevik and Ho-Kyeong Ra and Viren Bajaj and Zeya Ahmad},
  title   = {UQLM: A Python Package for Uncertainty Quantification in Large Language Models},
  journal = {Journal of Machine Learning Research},
  year    = {2026},
  volume  = {27},
  number  = {13},
  pages   = {1--10},
  url     = {http://jmlr.org/papers/v27/25-1557.html}
}

@inproceedings{
jiang2024graphbaseduncertaintymetricslongform,
title={Graph-based Uncertainty Metrics for Long-form Language Model Generations},
author={Mingjian Jiang and Yangjun Ruan and Prasanna Sattigeri and Salim Roukos and Tatsunori Hashimoto},
booktitle={The Thirty-eighth Annual Conference on Neural Information Processing Systems},
year={2024},
url={https://openreview.net/forum?id=YgJPQW0lkO}
}

@article{shorinwa2024surveyuncertaintyquantificationlarge,
author = {Shorinwa, Ola and Mei, Zhiting and Lidard, Justin and Ren, Allen Z. and Majumdar, Anirudha},
title = {A Survey on Uncertainty Quantification of Large Language Models: Taxonomy, Open Research Challenges, and Future Directions},
year = {2025},
issue_date = {February 2026},
publisher = {Association for Computing Machinery},
address = {New York, NY, USA},
volume = {58},
number = {3},
issn = {0360-0300},
url = {https://doi.org/10.1145/3744238},
doi = {10.1145/3744238},
journal = {ACM Comput. Surv.},
month = sep,
articleno = {63},
numpages = {38}
}

@article{huang2023surveyhallucinationlargelanguage,
author = {Huang, Lei and Yu, Weijiang and Ma, Weitao and Zhong, Weihong and Feng, Zhangyin and Wang, Haotian and Chen, Qianglong and Peng, Weihua and Feng, Xiaocheng and Qin, Bing and Liu, Ting},
title = {A Survey on Hallucination in Large Language Models: Principles, Taxonomy, Challenges, and Open Questions},
year = {2025},
issue_date = {March 2025},
publisher = {Association for Computing Machinery},
address = {New York, NY, USA},
volume = {43},
number = {2},
issn = {1046-8188},
url = {https://doi.org/10.1145/3703155},
doi = {10.1145/3703155},
journal = {ACM Trans. Inf. Syst.},
month = jan,
articleno = {42},
numpages = {55}
}

@inproceedings{chen2023quantifyinguncertaintyanswerslanguage,
    title = "Quantifying Uncertainty in Answers from any Language Model and Enhancing their Trustworthiness",
    author = "Chen, Jiuhai  and
      Mueller, Jonas",
    editor = "Ku, Lun-Wei  and
      Martins, Andre  and
      Srikumar, Vivek",
    booktitle = "Proceedings of the 62nd Annual Meeting of the Association for Computational Linguistics (Volume 1: Long Papers)",
    month = aug,
    year = "2024",
    address = "Bangkok, Thailand",
    publisher = "Association for Computational Linguistics",
    url = "https://aclanthology.org/2024.acl-long.283/",
    doi = "10.18653/v1/2024.acl-long.283",
    pages = "5186--5200"
}

@Article{Farquhar2024,
author={Farquhar, Sebastian
and Kossen, Jannik
and Kuhn, Lorenz
and Gal, Yarin},
title={Detecting hallucinations in large language models using semantic entropy},
journal={Nature},
year={2024},
month={Jun},
day={01},
volume={630},
number={8017},
pages={625-630},
issn={1476-4687},
doi={10.1038/s41586-024-07421-0},
url={https://doi.org/10.1038/s41586-024-07421-0}
}

@inproceedings{zhang2024luqlongtextuncertaintyquantification,
    title = "{LUQ}: Long-text Uncertainty Quantification for {LLM}s",
    author = "Zhang, Caiqi  and
      Liu, Fangyu  and
      Basaldella, Marco  and
      Collier, Nigel",
    editor = "Al-Onaizan, Yaser  and
      Bansal, Mohit  and
      Chen, Yun-Nung",
    booktitle = "Proceedings of the 2024 Conference on Empirical Methods in Natural Language Processing",
    month = nov,
    year = "2024",
    address = "Miami, Florida, USA",
    publisher = "Association for Computational Linguistics",
    url = "https://aclanthology.org/2024.emnlp-main.299/",
    doi = "10.18653/v1/2024.emnlp-main.299",
    pages = "5244--5262"
}

@inproceedings{manakul2023selfcheckgptzeroresourceblackboxhallucination,
    title = "{S}elf{C}heck{GPT}: Zero-Resource Black-Box Hallucination Detection for Generative Large Language Models",
    author = "Manakul, Potsawee  and
      Liusie, Adian  and
      Gales, Mark",
    editor = "Bouamor, Houda  and
      Pino, Juan  and
      Bali, Kalika",
    booktitle = "Proceedings of the 2023 Conference on Empirical Methods in Natural Language Processing",
    month = dec,
    year = "2023",
    address = "Singapore",
    publisher = "Association for Computational Linguistics",
    url = "https://aclanthology.org/2023.emnlp-main.557/",
    doi = "10.18653/v1/2023.emnlp-main.557",
    pages = "9004--9017"
}

@article{
lin2024generatingconfidenceuncertaintyquantification,
title={Generating with Confidence: Uncertainty Quantification for Black-box Large Language Models},
author={Zhen Lin and Shubhendu Trivedi and Jimeng Sun},
journal={Transactions on Machine Learning Research},
issn={2835-8856},
year={2024},
url={https://openreview.net/forum?id=DWkJCSxKU5},
note={}
}

@inproceedings{
kuhn2023semanticuncertaintylinguisticinvariances,
title={Semantic Uncertainty: Linguistic Invariances for Uncertainty Estimation in Natural Language Generation},
author={Lorenz Kuhn and Yarin Gal and Sebastian Farquhar},
booktitle={The Eleventh International Conference on Learning Representations },
year={2023},
url={https://openreview.net/forum?id=VD-AYtP0dve}
}

@inproceedings{
qiu2024semanticdensityuncertaintyquantification,
title={Semantic Density: Uncertainty Quantification for Large Language Models through Confidence Measurement in Semantic Space},
author={Xin Qiu and Risto Miikkulainen},
booktitle={The Thirty-eighth Annual Conference on Neural Information Processing Systems},
year={2024},
url={https://openreview.net/forum?id=LOH6qzI7T6}
}

@inproceedings{
cole2023selectivelyansweringambiguousquestions,
title={Selectively Answering Ambiguous Questions},
author={Jeremy R. Cole and Michael JQ Zhang and Daniel Gillick and Julian Martin Eisenschlos and Bhuwan Dhingra and Jacob Eisenstein},
booktitle={The 2023 Conference on Empirical Methods in Natural Language Processing},
year={2023},
url={https://openreview.net/forum?id=x2W2dKdNI8}
}

@inproceedings{
zhang2020bertscoreevaluatingtextgeneration,
title={BERTScore: Evaluating Text Generation with BERT},
author={Tianyi Zhang* and Varsha Kishore* and Felix Wu* and Kilian Q. Weinberger and Yoav Artzi},
booktitle={International Conference on Learning Representations},
year={2020},
url={https://openreview.net/forum?id=SkeHuCVFDr}
}

@inproceedings{
xiong2024llmsexpressuncertaintyempirical,
title={Can {LLM}s Express Their Uncertainty? An Empirical Evaluation of Confidence Elicitation in {LLM}s},
author={Miao Xiong and Zhiyuan Hu and Xinyang Lu and YIFEI LI and Jie Fu and Junxian He and Bryan Hooi},
booktitle={The Twelfth International Conference on Learning Representations},
year={2024},
url={https://openreview.net/forum?id=gjeQKFxFpZ}
}

@misc{kadavath2022languagemodelsmostlyknow,
      title={Language Models (Mostly) Know What They Know}, 
      author={Saurav Kadavath and Tom Conerly and Amanda Askell and Tom Henighan and Dawn Drain and Ethan Perez and Nicholas Schiefer and Zac Hatfield-Dodds and Nova DasSarma and Eli Tran-Johnson and Scott Johnston and Sheer El-Showk and Andy Jones and Nelson Elhage and Tristan Hume and Anna Chen and Yuntao Bai and Sam Bowman and Stanislav Fort and Deep Ganguli and Danny Hernandez and Josh Jacobson and Jackson Kernion and Shauna Kravec and Liane Lovitt and Kamal Ndousse and Catherine Olsson and Sam Ringer and Dario Amodei and Tom Brown and Jack Clark and Nicholas Joseph and Ben Mann and Sam McCandlish and Chris Olah and Jared Kaplan},
      year={2022},
      eprint={2207.05221},
      archivePrefix={arXiv},
      primaryClass={cs.CL},
      url={https://arxiv.org/abs/2207.05221}, 
}

@misc{verga2024replacingjudgesjuriesevaluating,
      title={Replacing Judges with Juries: Evaluating LLM Generations with a Panel of Diverse Models}, 
      author={Pat Verga and Sebastian Hofstatter and Sophia Althammer and Yixuan Su and Aleksandra Piktus and Arkady Arkhangorodsky and Minjie Xu and Naomi White and Patrick Lewis},
      year={2024},
      eprint={2404.18796},
      archivePrefix={arXiv},
      primaryClass={cs.CL},
      url={https://arxiv.org/abs/2404.18796}, 
}

@inproceedings{fadeeva2024factcheckingoutputlargelanguage,
    title = "Fact-Checking the Output of Large Language Models via Token-Level Uncertainty Quantification",
    author = "Fadeeva, Ekaterina  and
      Rubashevskii, Aleksandr  and
      Shelmanov, Artem  and
      Petrakov, Sergey  and
      Li, Haonan  and
      Mubarak, Hamdy  and
      Tsymbalov, Evgenii  and
      Kuzmin, Gleb  and
      Panchenko, Alexander  and
      Baldwin, Timothy  and
      Nakov, Preslav  and
      Panov, Maxim",
    editor = "Ku, Lun-Wei  and
      Martins, Andre  and
      Srikumar, Vivek",
    booktitle = "Findings of the Association for Computational Linguistics: ACL 2024",
    month = aug,
    year = "2024",
    address = "Bangkok, Thailand",
    publisher = "Association for Computational Linguistics",
    url = "https://aclanthology.org/2024.findings-acl.558/",
    doi = "10.18653/v1/2024.findings-acl.558",
    pages = "9367--9385"
}

@inproceedings{
malinin2021uncertaintyestimationautoregressivestructured,
title={Uncertainty Estimation in Autoregressive Structured Prediction},
author={Andrey Malinin and Mark Gales},
booktitle={International Conference on Learning Representations},
year={2021},
url={https://openreview.net/forum?id=jN5y-zb5Q7m}
}

\appendix
\section{Scorer Definitions}
\label{sec:scorer_details}
We provide formal definitions of all UQ scorers used as ensemble inputs. All scorers are constructed to output values in [0, 1] by design, so no additional preprocessing or calibration is applied before combination. Let $y$ denote the original response to prompt $x$, with tokenization $\{t_1, \ldots, t_L\}$ where $L$ is the number of tokens and $p_j$ is the probability assigned to token $t_j$. For sampling-based methods, $\tilde{\mathbf{y}} = \{\tilde{y}_1, \ldots, \tilde{y}_m\}$ denotes $m$ candidate responses generated from the same prompt using stochastic decoding.

\subsection{White-Box Single-Generation Scorers}

These scorers derive confidence from token-level probabilities of a single generation.

\paragraph{Sequence Probability (SP).} The joint probability of all tokens \citep{Vashurin_2025}:
$$\text{SP}(y) = \prod_{j=1}^{L} p_j$$

\paragraph{Length-Normalized Sequence Probability (LNSP).} The geometric mean of token probabilities, correcting for sequence length \citep{malinin2021uncertaintyestimationautoregressivestructured}:
$$\text{LNSP}(y) = \left(\prod_{j=1}^{L} p_j\right)^{1/L}$$

\paragraph{Minimum Token Probability (MTP).} The minimum token probability across the response \citep{manakul2023selfcheckgptzeroresourceblackboxhallucination}:
$$\text{MTP}(y) = \min_{j \in \{1,\ldots,L\}} p_j$$

The following scorers require access to the top-$K$ logprobs per token. Let $\{p_{j,1}, \ldots, p_{j,K}\}$ denote the top-$K$ token probabilities at position $j$, ordered by decreasing probability.

\paragraph{Probability Margin (PM).} The average gap between the top two token probabilities \citep{farr2024redctsystemsdesignmethodology}:
$$\text{PM}(y) = \frac{1}{L} \sum_{j=1}^{L} (p_{j,1} - p_{j,2})$$

\paragraph{Average Token Negentropy (ATN@$K$).} The mean normalized negentropy across token positions \citep{scalena2025eagerentropyawaregenerationadaptive, manakul2023selfcheckgptzeroresourceblackboxhallucination}. Top-$K$ token entropy at position $j$ is $\text{TE@}K(t_j) = -\sum_{k=1}^{K} p_{j,k} \log p_{j,k}$. The negentropy transformation normalizes to $[0,1]$:
$$\text{TN@}K(t_j) = 1 - \frac{\text{TE@}K(t_j)}{\log K} $$
$$\text{ATN@}K(y) = \frac{1}{L} \sum_{j=1}^{L} \text{TN@}K(t_j)$$

\paragraph{Minimum Token Negentropy (MTN@$K$).} The minimum token negentropy across positions \citep{scalena2025eagerentropyawaregenerationadaptive, manakul2023selfcheckgptzeroresourceblackboxhallucination}:
$$\text{MTN@}K(y) = \min_{j \in \{1,\ldots,L\}} \text{TN@}K(t_j)$$

\subsection{Black-Box Sampling-Based Scorers}

These scorers generate $m$ candidate responses and measure consistency with the original response. All scorers in this subsection require only text outputs (no token probabilities).

\paragraph{Exact Match Rate (EMR).} The proportion of candidates identical to the original \citep{cole2023selectivelyansweringambiguousquestions}:
$$\text{EMR}(y; \tilde{\mathbf{y}}) = \frac{1}{m} \sum_{j=1}^{m} \mathbb{I}(y = \tilde{y}_j)$$

\paragraph{Non-Contradiction Probability (NCP).} The mean bidirectional non-contradiction probability from an NLI model \citep{chen2023quantifyinguncertaintyanswerslanguage}:
$$\text{NCP}(y; \tilde{\mathbf{y}}) = 1 - \frac{1}{m} \sum_{j=1}^{m} \frac{p_c(y, \tilde{y}_j) + p_c(\tilde{y}_j, y)}{2}$$
where $p_c(\cdot, \cdot)$ denotes the NLI contradiction probability. We use \texttt{microsoft/deberta-large-mnli} for all NLI-based scorers.

\paragraph{Entailment Probability (EP).} The mean bidirectional entailment probability \citep{chen2023quantifyinguncertaintyanswerslanguage, lin2024generatingconfidenceuncertaintyquantification}:
$$\text{EP}(y; \tilde{\mathbf{y}}) = \frac{1}{m} \sum_{j=1}^{m} \frac{p_e(y, \tilde{y}_j) + p_e(\tilde{y}_j, y)}{2}$$
where $p_e(\cdot, \cdot)$ denotes the NLI entailment probability.

\paragraph{BERTScore Consistency (BSC).} The average F1 BERTScore between the original and each candidate \citep{zhang2020bertscoreevaluatingtextgeneration, manakul2023selfcheckgptzeroresourceblackboxhallucination}:
$$\text{BSC}(y; \tilde{\mathbf{y}}) = \frac{1}{m} \sum_{j=1}^{m} \text{BertF1}(y, \tilde{y}_j)$$

\paragraph{Normalized Cosine Similarity (NCS).} The average cosine similarity using a sentence embedding model $V: \mathcal{Y} \to \mathbb{R}^d$, normalized to $[0,1]$ \citep{bouchard2025uncertainty}:
$$\text{NCS}(y; \tilde{\mathbf{y}}) = \frac{1}{2} + \frac{1}{2m} \sum_{j=1}^{m} \frac{V(y) \cdot V(\tilde{y}_j)}{\|V(y)\| \cdot \|V(\tilde{y}_j)\|}.$$
We use \texttt{sentence-transformers/all-MiniLM-L6-v2} for natural language embeddings and \texttt{jinaai/jina-embeddings-v2-base-code} for code embeddings.

\paragraph{Normalized Semantic Negentropy (NSN).} Responses are clustered by mutual entailment via an NLI model. Semantic entropy is computed over the cluster distribution and normalized to a confidence score in $[0,1]$ \citep{kuhn2023semanticuncertaintylinguisticinvariances, Farquhar2024, bouchard2025uncertainty}:
$$\text{SE}(y; \tilde{\mathbf{y}}) = -\sum_{C \in \mathcal{C}} P(C) \log P(C)$$

$$\text{NSN}(y; \tilde{\mathbf{y}}) = 1 - \frac{\text{SE}(y; \tilde{\mathbf{y}})}{\log(m+1)}$$
where $\mathcal{C}$ is the set of clusters over $\{y\} \cup \tilde{\mathbf{y}}$ and $P(C)$ is the proportion of responses in cluster $C$.

\paragraph{Semantic Sets Confidence (SSC).} The number of unique semantic clusters $|\mathcal{C}|$, normalized to $[0,1]$ \citep{lin2024generatingconfidenceuncertaintyquantification}:
$$\text{SSC}(y; \tilde{\mathbf{y}}) = \frac{m + 1 - |\mathcal{C}|}{m}$$

\subsection{Hybrid Scorers}

These scorers combine token probabilities with sampling-based consistency signals.

\paragraph{Monte Carlo Sequence Probability (MCSP).} The average length-normalized sequence probability across all sampled responses \citep{kuhn2023semanticuncertaintylinguisticinvariances}:
$$\text{MCSP}(\tilde{\mathbf{y}}) = \frac{1}{m+1} \sum_{i=0}^{m} \text{LNSP}(y_i)$$
where $y_0 = y$ is the original response.

\paragraph{Consistency and Confidence Approach (CoCoA).} The product of the original response's length-normalized sequence probability and its normalized cosine similarity with sampled responses \citep{vashurin2025uncertaintyquantificationllmsminimum}:
$$\text{CoCoA}(y; \tilde{\mathbf{y}}) = \text{LNSP}(y) \cdot \text{NCS}(y; \tilde{\mathbf{y}})$$

\subsection{Reflexive Scorers}

These scorers prompt the LLM to evaluate its own output. Both are applicable across all generation regimes (short-form, long-form, and code generation).

\paragraph{P(True).} The model is presented with a question-response concatenation and asked to classify it as ``True'' or ``False.'' Confidence is the token probability assigned to ``True'' \citep{kadavath2022languagemodelsmostlyknow}:
$$\text{P(True)}(y; x) = p(\text{``True''} \mid x, y)$$

\paragraph{Verbalized Confidence (VC).} The model is prompted to express its confidence as a numerical score on a scale from 0 to 1, without requiring access to token probabilities \citep{tian2023justaskcalibrationstrategies}. We implement a six-level likert scale that maps to numerical values $\{0, 0.2,...,1.0\}$ \citep{xiong2024llmsexpressuncertaintyempirical}.

\subsection{Code-Specific Scorers}

For code generation, we adapt several sampling-based scorers by replacing NLI-based semantic equivalence with LLM-based functional equivalence assessment, which judges whether two code snippets produce identical outputs for all valid inputs.

\paragraph{Functional Equivalence Rate (FER).} The proportion of sampled responses judged functionally equivalent to the original \citep{bouchard2026functionalentropy}:
$$\text{FER}(y; \tilde{\mathbf{y}}) = \frac{1}{m} \sum_{j=1}^{m} \mathbb{I}[y \equiv \tilde{y}_j]$$

\paragraph{Functional Entropy (FE).} A code-specific analogue of semantic entropy, using functional equivalence for clustering. Normalized to $[0,1]$ \citep{bouchard2026functionalentropy}:
$$\text{FE}(y; \tilde{\mathbf{y}}) = -\sum_{C \in \mathcal{C}} P(C) \log P(C)$$ 
$$\text{NFN}(y; \tilde{\mathbf{y}}) = 1 - \frac{\text{FE}(y; \tilde{\mathbf{y}})}{\log(m+1)}$$

\paragraph{Functional Sets Confidence (FSC).} The normalized count of unique functional clusters \citep{bouchard2026functionalentropy}:
$$\text{FSC}(y; \tilde{\mathbf{y}}) = \frac{m + 1 - |\mathcal{C}|}{m}$$

\paragraph{CodeBLEU Consistency (CBC).} The average CodeBLEU score between the original and each candidate, capturing structural and syntactic similarity via n-gram match, syntax trees, and data-flow analysis \citep{ren2020codebleumethodautomaticevaluation, sharma2025assessingcorrectnessllmbasedcode}:
$$\text{CBC}(y; \tilde{\mathbf{y}}) = \frac{1}{m} \sum_{j=1}^{m} \text{CodeBLEU}(y, \tilde{y}_j)$$

\subsection{Long-Form Claim-Level Scorers}

For long-form generation, scorers operate at the claim level following a three-stage pipeline: (1) decompose the response into claims, (2) score each claim, and (3) aggregate to a response-level confidence. In this work, the ensemble operates at the claim level (stage 2), so we describe the claim-level scoring below. We employ graph-based scorers proposed by \citet{jiang2024graphbaseduncertaintymetricslongform} and extended by \citet{bouchard2026finegraineduncertaintyquantificationlongform}, which decompose both original and sampled responses into claims, obtain the union of unique claims $\mathbf{s}$ across all responses, and construct a bipartite graph $G$ with node set $V = \mathbf{s} \cup \mathbf{y}$, where an edge exists between claim $s$ and response $y$ if and only if $s$ is entailed in $y$.

\paragraph{Degree Centrality.} The average entailment probability across responses:
$$\text{DC}(s) = \frac{1}{m} \sum_{j=1}^{m} P(\text{entail} \mid y_j, s).$$
Note that degree centrality scores each claim against the sampled responses only and does not compare claims across pairs of sampled responses. This quantity coincides with the per-claim score underlying LUQ-Atomic \citep{zhang2024luqlongtextuncertaintyquantification}, up to the choice of entailment probability estimator.

\paragraph{Betweenness Centrality.} The proportion of shortest paths between node pairs passing through $s$, normalized by the maximum possible value $B_{\text{max}}$:
$$\text{BC}(s) = \frac{1}{B_{\text{max}}} \sum_{u \neq v \neq s} \frac{\sigma_{uv}(s)}{\sigma_{uv}}$$
where $\sigma_{uv}$ is the number of shortest paths between $u$ and $v$, and $\sigma_{uv}(s)$ is the number passing through $s$.

\paragraph{Closeness Centrality.} The inverse sum of distances to all other nodes, normalized by the minimum possible distance:
$$\text{CC}(s) = \frac{m + 2(|\mathbf{s}| - 1)}{\sum_{v \neq s} \text{dist}(s, v)}$$

\paragraph{Harmonic Centrality.} The sum of inverse distances, normalized by the maximum possible value $H_{\text{max}} = m + \frac{|\mathbf{s}| - 1}{2}$:
$$\text{HC}(s) = \frac{1}{H_{\text{max}}} \sum_{v \neq s} \frac{1}{\text{dist}(s, v)}$$

\paragraph{Laplacian Centrality.} The proportional drop in Laplacian energy from removing $s$:
$$\text{LC}(s) = \frac{E_L(G) - E_L(G_{-s})}{E_L(G)}$$
where $E_L(G) = \sum_i \lambda_i^2$ and $\lambda_i$ are the eigenvalues of $G$'s Laplacian matrix.

\paragraph{PageRank.} The stationary distribution probability of a random walk with restart probability $(1-d)$:
$$\text{PR}(s) = \frac{1-d}{|V|} + d \sum_{v \in N(s)} \frac{\text{PR}(v)}{|N(v)|}$$
where $N(s)$ is the set of neighbors of $s$.

\section{Computational Cost}
\label{sec:cost}
\begin{table*}[h]
\centering
\small
\begin{tabular}{lcccc}
\toprule
\textbf{Scorer Family} & \textbf{Regime} & \textbf{Orig.\ LLM} & \textbf{Aux.\ LLM} & \textbf{Sem.\ Comp.} \\
\midrule
Single-gen.\ white-box & SF, CG & 0 & 0 & 0 \\
Sampling-based & SF, CG & $m$ & 0 & $m$--$\binom{m+1}{2}$ \\
Reflexive & All & 1 & 0 & 0 \\
Graph-based & LF & $m$ & $2m + 1$ & $m \cdot N_{\text{claims}}$ \\
\bottomrule
\end{tabular}
\caption{Per-instance computational cost by scorer family. SF = short-form, CG = code generation, LF = long-form. ``Orig.\ LLM'' = additional generations from the model under evaluation. ``Aux.\ LLM'' = generations from a separate model. ``Sem.\ Comp.'' = pairwise semantic comparisons. $m$ = number of sampled responses; $N_{\text{claims}}$ = number of unique claims across sampled responses.}
\label{tab:cost}
\end{table*}

Table~\ref{tab:cost} summarizes the per-instance computational cost of each scorer family beyond the initial response generation. Costs are expressed in terms of additional generations from the original LLM, auxiliary generations from a separate model (e.g., for claim decomposition or entailment grading), and semantic comparisons (e.g., NLI inference, embedding similarity, CodeBLEU, or LLM-based equivalence checks).

Single-generation white-box scorers incur no cost beyond extracting token probabilities from the original forward pass. Sampling-based scorers require $m$ additional generations and between $m$ (for pairwise comparison against the original only) and $\binom{m+1}{2}$ (for all-pairs clustering, as in semantic entropy) semantic comparisons. Reflexive scorers require one additional generation from the same LLM in which the model evaluates its own output. Graph-based scorers, used only in the long-form setting, are the most expensive: they require $m$ sampled responses, $2m + 1$ auxiliary LLM calls for claim decomposition ($m + 1$ responses decomposed into claims) and claim merging ($m$ merge operations), and $m \cdot N_{\text{claims}}$ entailment checks to construct the claim-response bipartite graph, where $N_{\text{claims}}$ denotes the number of unique claims across sampled responses.

The total cost of the ensemble for a given regime is the sum of costs across the applicable scorer families. For short-form and code generation, this is the combined cost of single-generation white-box, sampling-based, and reflexive scorers. For long-form, this is the combined cost of graph-based and reflexive scorers.
 \begin{table*}[t]
    \centering
    \tiny
    \begin{tabular}{lcccccccc}
        \toprule
        & \multicolumn{5}{c}{\textbf{Short-Form}} & \textbf{Code} & \multicolumn{2}{c}{\textbf{Long-Form}} \\
        \cmidrule(lr){2-6} \cmidrule(lr){7-7} \cmidrule(lr){8-9}
        \textbf{LLM} & BigMath & OpenR1 & DROP  & Hotpot & SimpleQA & Python & Rivers & Mushrooms \\
        \midrule
        Gemini-2.5-Flash & 0.94 & 0.89 & 0.84  & 0.93 & 0.31 & 0.87 & 0.50 & 0.55 \\
        Gemini-2.5-Pro   & 0.93 & 0.92 & 0.84  & 0.94 & 0.54 & 0.91 & 0.48 & 0.56 \\
        GPT-4o           & 0.40 & 0.25 & 0.78  & 0.93 & 0.27 & 0.56 & 0.54 & 0.61 \\
        GPT-4o-mini      & 0.45 & 0.24 & 0.75  & 0.89 & 0.08 & 0.52 & 0.47 & 0.52 \\
        \bottomrule
    \end{tabular}%
    \caption{LLM accuracy across evaluation datasets. Accuracy is computed as the average over binary correctness labels, which serve as ground truth for evaluating uncertainty quantification methods.}

        \label{tab:llm_accuracy}
\end{table*}

\section{Long-Form Scoring and Grading}
\label{sec:longform_datasets}
 \paragraph{Dataset Construction.}
We construct two long-form QA datasets following the FactScore \citep{min2023factscorefinegrainedatomicevaluation} protocol. For each dataset, entities are drawn from Wikipedia: 84 edible mushroom species and 500 rivers. Each entity is paired with the prompt ``Write a paragraph detailing some facts about \{entity\},'' where \{entity\} is the mushroom species or river name. Wikipedia articles are retrieved via the Wikipedia API to serve as reference texts. The full entity lists and reproducibility code are provided in the supplemental materials.

\paragraph{Long-Form Scoring Pipeline.}
Long-form scoring operates as follows. First, each response is decomposed into atomic claims using the prompt template from \citet{zhang2025atomiccalibrationllmslongform}. Second, the union of unique claims across all responses is obtained via sequential claim merging, following \citet{jiang2024graphbaseduncertaintymetricslongform}: each new claim is compared against the existing set and merged with a matching claim if one exists, or appended as a new entry otherwise. This produces a deduplicated claim set $s$ over which the graph-based scorers (Section~\ref{sec:scorers}) operate.  Graph-based scorers measure how consistently each claim is entailed across sampled responses via centrality metrics on the claim-response bipartite graph rather than scoring claims against the original query. Reflexive scorers, by contrast, are conditioned on the original question and evaluate each claim in that context. The ensemble operates at the claim level: classification and evaluation both use claim-level labels $h(c)$, and no response-level aggregation is required.

\paragraph{Long-Form (Claim-Level) Grading.}
Each claim is classified as objective or subjective following \citet{zhang2024luqlongtextuncertaintyquantification}; only objective claims are retained for evaluation, as subjective claims cannot be definitively verified against a reference. Each retained objective claim is then graded against the entity's complete Wikipedia article using the FactScore protocol \citep{min2023factscorefinegrainedatomicevaluation, zhang2025atomiccalibrationllmslongform, zhang2024luqlongtextuncertaintyquantification, jiang2024graphbaseduncertaintymetricslongform}, producing the binary labels $h(c) \in \{0, 1\}$ used for both training and evaluation. Gemini-2.5-Flash is used for claim decomposition, claim merging, objectivity classification, and grading, chosen for its strong performance at low cost.

\section{Grading Validation}
\label{sec:grading_validation}
To assess the reliability of LLM-based grading, two human annotators independently labeled a stratified sample of 400 short-form responses. For each of the five short-form datasets, we sampled 20 responses marked correct and 20 marked incorrect by the Gemini-2.5-Flash grader, across two generator LLMs (GPT-4o and Gemini-2.5-Flash), yielding 80 responses per dataset. Annotators compared each generated answer against the reference answer provided in the original dataset, without access to the grader's label.

Table~\ref{tab:overall_agreement} reports overall pairwise agreement. All three comparisons yield Cohen's $\kappa \geq 0.93$, indicating near-perfect agreement. Notably, the LLM grader agrees with each annotator at least as strongly as the annotators agree with each other ($\kappa = 0.97$ and $0.93$ vs.\ $0.95$), confirming that grading noise introduced by the LLM is no larger than inherent human disagreement.

Table~\ref{tab:agreement_by_dataset} reports agreement broken down by dataset. Agreement is highest on math datasets ($\kappa \geq 0.95$), where correctness is unambiguous, and lowest on HotpotQA (human $\kappa = 0.85$). Even on HotpotQA, the LLM grader agrees with Annotator 1 more than the annotators agree with each other ($\kappa = 0.97$ vs.\ $0.85$), suggesting that disagreements stem from reference answer ambiguity rather than grader error.

Table~\ref{tab:agreement_by_generator} reports agreement broken down by generator LLM to test whether the Gemini-2.5-Flash grader favors its own responses. Human-human agreement is identical across generators ($\kappa = 0.95$), and LLM-human agreement is comparable (LLM vs.\ Annotator 1: $\kappa = 0.97$ for Gemini-2.5-Flash vs.\ $0.98$ for GPT-4o; LLM vs.\ Annotator 2: $\kappa = 0.92$ vs.\ $0.93$). These results provide no evidence of self-grading bias.

\begin{table}[t]
\centering
\tiny
\begin{tabular}{lcc}
\toprule
\textbf{Comparison} & \textbf{\% Agreement} & \textbf{Cohen's $\kappa$} \\
\midrule
Annotator 1 vs.\ Annotator 2 & 97.5\% & 0.95 \\
LLM Grader vs.\ Annotator 1 & 98.8\% & 0.97 \\
LLM Grader vs.\ Annotator 2 & 96.2\% & 0.93 \\
\bottomrule
\end{tabular}
\caption{Overall pairwise agreement for grading validation ($n = 400$).}
\label{tab:overall_agreement}
\end{table}

\begin{table}[t]
\centering
\tiny
\begin{tabular}{lccccccc}
\toprule
& & \multicolumn{2}{c}{\textbf{Human-Human}} & \multicolumn{2}{c}{\textbf{LLM vs.\ Ann.\ 1}} & \multicolumn{2}{c}{\textbf{LLM vs.\ Ann.\ 2}} \\
\cmidrule(lr){3-4} \cmidrule(lr){5-6} \cmidrule(lr){7-8}
\textbf{Dataset} & \textbf{$n$} & \textbf{\%} & \textbf{$\kappa$} & \textbf{\%} & \textbf{$\kappa$} & \textbf{\%} & \textbf{$\kappa$} \\
\midrule
BigMath & 80 & 98.8\% & 0.97 & 98.8\% & 0.97 & 97.5\% & 0.95 \\
DROP & 80 & 97.5\% & 0.95 & 98.8\% & 0.97 & 96.2\% & 0.93 \\
HotpotQA & 80 & 92.5\% & 0.85 & 98.8\% & 0.97 & 91.2\% & 0.82 \\
OpenR1 & 80 & 98.8\% & 0.97 & 100.0\% & 1.00 & 98.8\% & 0.97 \\
SimpleQA & 80 & 100.0\% & 1.00 & 97.5\% & 0.95 & 97.5\% & 0.95 \\
\bottomrule
\end{tabular}
\caption{Agreement by dataset.}
\label{tab:agreement_by_dataset}
\end{table}

\begin{table}[t]
\centering
\tiny
\begin{tabular}{lccccccc}
\toprule
& & \multicolumn{2}{c}{\textbf{Human-Human}} & \multicolumn{2}{c}{\textbf{LLM vs.\ A1}} & \multicolumn{2}{c}{\textbf{LLM vs.\ A2}} \\
\cmidrule(lr){3-4} \cmidrule(lr){5-6} \cmidrule(lr){7-8}
\textbf{Orig. LLM } & \textbf{$n$} & \textbf{\%} & \textbf{$\kappa$} & \textbf{\%} & \textbf{$\kappa$} & \textbf{\%} & \textbf{$\kappa$} \\
\midrule
Gem-Flash & 200 & 97.5\% & 0.95 & 98.5\% & 0.97 & 96.0\% & 0.92 \\
GPT-4o & 200 & 97.5\% & 0.95 & 99.0\% & 0.98 & 96.5\% & 0.93 \\
\bottomrule
\end{tabular}
\caption{Agreement by generator LLM, testing for self-grading bias. A1 and A2 respectively refer to the two annotators.}
\label{tab:agreement_by_generator}
\end{table}

\section{Hyperparameters}
\label{sec:hyperparameters}

All combination strategies use 5-fold cross-validation on the training fold for hyperparameter selection, optimizing AUROC. We use \texttt{uqlm} \citep{bouchard2025uqlmpythonpackageuncertainty} for the weighted average method and \texttt{scikit-learn} \citep{JMLR:v12:pedregosa11a} for the other three classifiers.

\paragraph{Weighted average.} Weights are constrained to $[0,1]$ and sum to 1. We optimize AUROC using Optuna \citep{Akiba_Optuna_A_next-generation_2019} with 1{,}000 trials per configuration.

\paragraph{Logistic regression.} We use elastic net regularization (\texttt{penalty='elasticnet'}, \texttt{solver='saga'}) with the following grid: regularization strength $C \in \{0.001, 0.01, 0.1, 1, 10, 100\}$ and $\ell_1$ ratio $\in \{0, 0.5, 1\}$, yielding 18 configurations.

\paragraph{Random forest.} We search over: \texttt{n\_estimators} $\in \{200, 500\}$, \texttt{max\_features} $\in \{\text{sqrt}, \text{log2}\}$, \texttt{max\_depth} $\in \{4, 6, 8\}$, \texttt{min\_samples\_split} $\in \{2, 5\}$, and \texttt{min\_samples\_leaf} $\in \{1, 2\}$, yielding 96 configurations.

\paragraph{Gradient boosted trees.} We search over: \texttt{n\_estimators} $\in \{50, 100, 200\}$, \texttt{learning\_rate} $\in \{0.01, 0.1, 0.2\}$, \texttt{max\_depth} $\in \{3, 4, 5\}$, \texttt{min\_samples\_split} $\in \{2, 4\}$, and \texttt{subsample} $\in \{0.8, 1.0\}$, yielding 108 configurations.

\section{Supplemental Figures and Tables}
\label{sec:appendix_tables}

\begin{figure*}[t]
    \centering
    
    \begin{subfigure}[b]{0.9\textwidth}
        \includegraphics[width=\textwidth]{figures/learning_curve_legend.png}
    \end{subfigure}
    \vspace{2mm}

    \begin{subfigure}[b]{0.9\textwidth}
        \includegraphics[width=\textwidth]{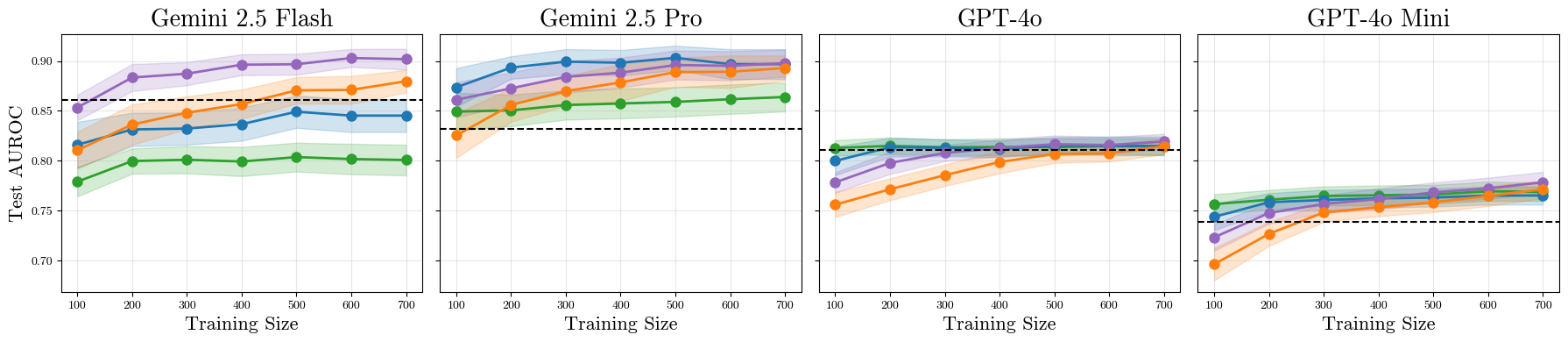}
        \caption{OpenR1-Math}
        \label{fig:openr1_learning}
    \end{subfigure}
    \vspace{2mm}

    \begin{subfigure}[b]{0.9\textwidth}
        \includegraphics[width=\textwidth]{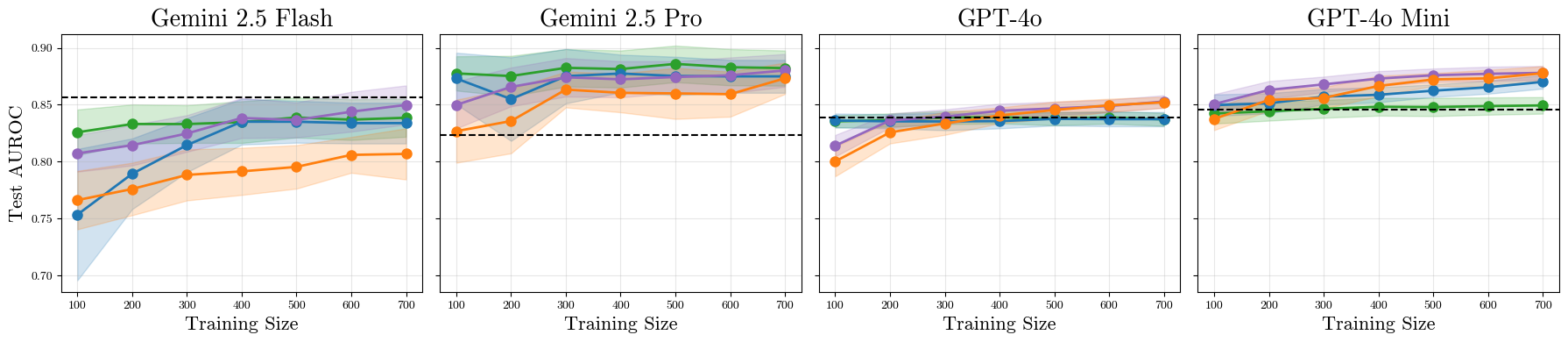}
        \caption{Big-Math}
        \label{fig:bigmath_learning}
    \end{subfigure}
    \vspace{2mm}

    \begin{subfigure}[b]{0.9\textwidth}
        \includegraphics[width=\textwidth]{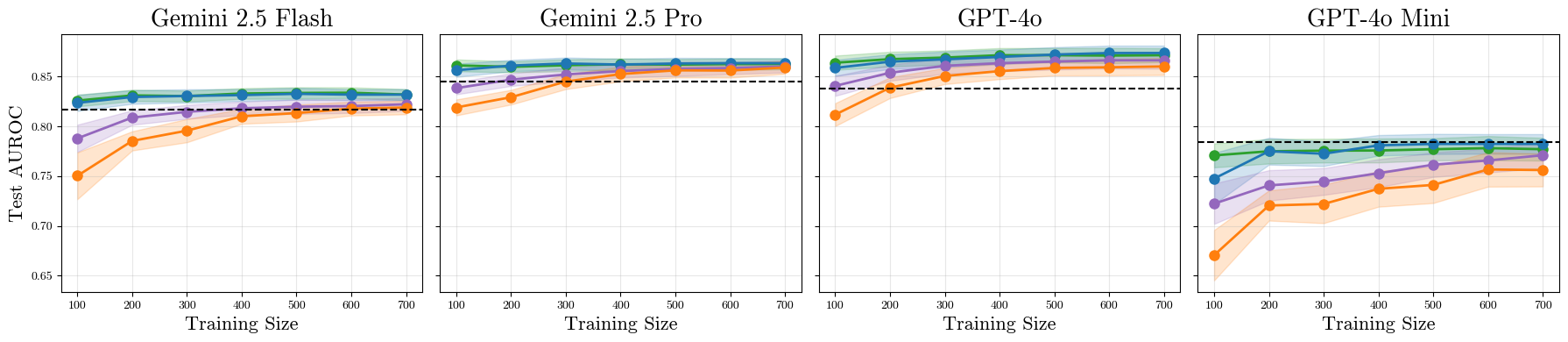}
        \caption{SimpleQA}
        \label{fig:simpleqa_learning}
    \end{subfigure}
    \vspace{2mm}

    \begin{subfigure}[b]{0.9\textwidth}
        \includegraphics[width=\textwidth]{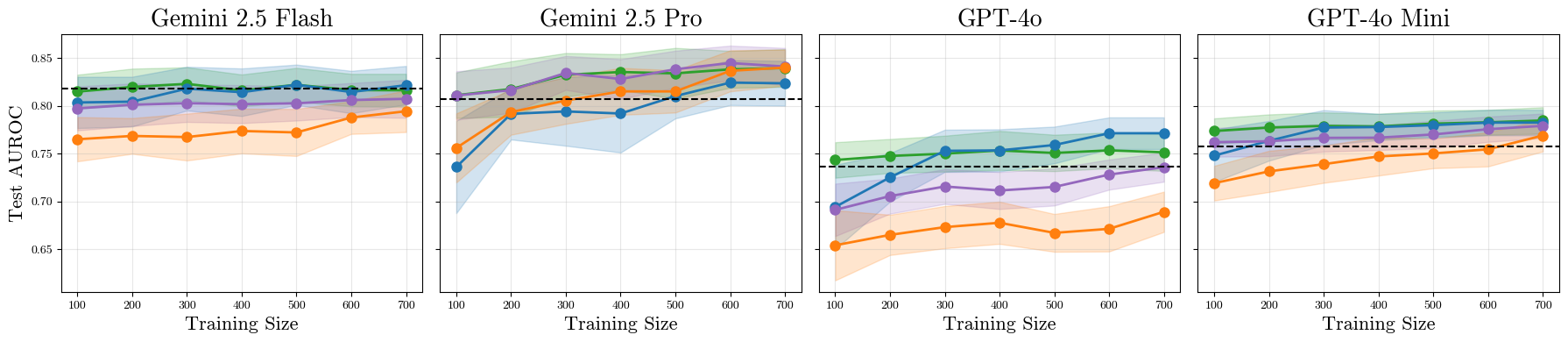}
        \caption{HotPotQA}
        \label{fig:hotpotqa_learning}
    \end{subfigure}
    \vspace{2mm}

    \begin{subfigure}[b]{0.9\textwidth}
        \includegraphics[width=\textwidth]{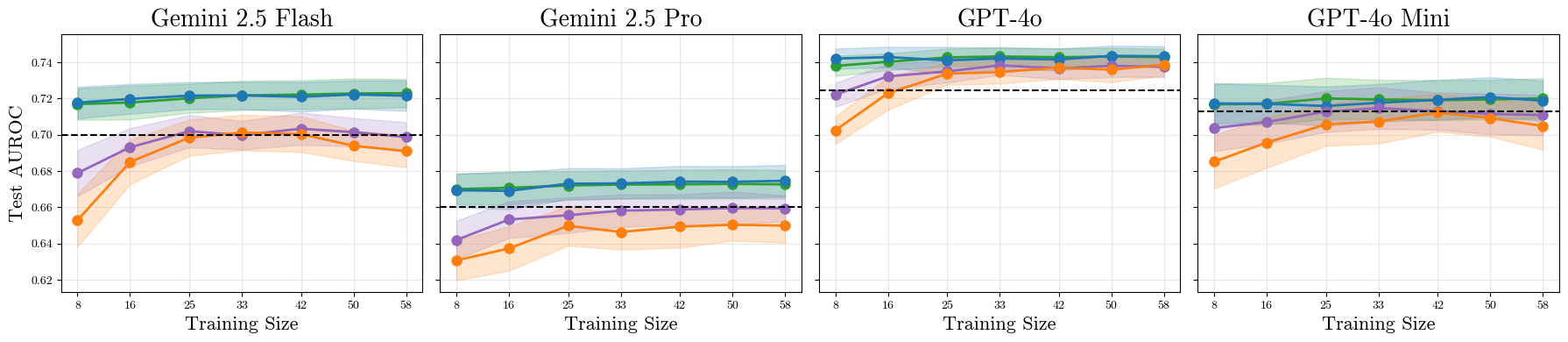}
        \caption{FactScore-Mushrooms}
        \label{fig:mushrooms_learning}
    \end{subfigure}

    \caption{Ensemble AUROC as a function of training sample size for remaining datasets. Lines show the four combination strategies with 95\% CIs over 25 splits. The dashed line is the best individual scorer, selected on the test set. Training sizes range from $0.1N$ to $0.7N$. See Figure~1 for representative datasets from each generation regime.}
    \label{fig:additional_learning_curves}
\end{figure*}

\begin{table*}[t]
\centering
\tiny
\setlength{\tabcolsep}{3pt}
\begin{tabular}{@{}lcccccccc@{}}
\toprule
 & \multicolumn{5}{c}{\textbf{Short-form QA}} & \textbf{Code} & \multicolumn{2}{c}{\textbf{Long-form}} \\
\cmidrule(lr){2-6} \cmidrule(lr){7-7} \cmidrule(lr){8-9}
\textbf{Scorer} & BigMath & OpenR1 & DROP & Hotpot & SimpleQA & LiveCodeBench & Mushroom-Fact & River-Fact \\
\midrule
\multicolumn{9}{l}{\textit{Single-generation white-box}} \\
\qquad Sequence prob. & 0.74 & 0.68 & 0.69 & 0.76 & 0.61 & 0.69 &  &  \\
\qquad Norm. sequence prob. & 0.74 & 0.70 & 0.66 & 0.76 & 0.60 &  &  &  \\
\qquad Min. token prob. & 0.74 & 0.67 & 0.68 & 0.77 & 0.60 & 0.77 &  &  \\
\qquad Probability margin & 0.74 & 0.70 & 0.66 & 0.75 & 0.60 & 0.66 &  &  \\
\qquad Mean token entropy & 0.74 & 0.70 & 0.66 & 0.75 & 0.60 & 0.69 &  &  \\
\qquad Min. token entropy & 0.74 & 0.68 & 0.68 & 0.76 & 0.60 & 0.77 &  &  \\
\multicolumn{9}{l}{\textit{Consistency-based black-box}} \\
\qquad Exact match rate & 0.77 & 0.61 & 0.67 & 0.80 & 0.79 &  &  &  \\
\qquad Non-contradiction prob. & 0.77 & 0.67 & 0.66 & 0.80 & 0.81 &  &  &  \\
\qquad Entailment prob. & 0.80 & 0.70 & 0.67 & 0.81 & 0.82 &  &  &  \\
\qquad BERTScore consistency & 0.74 & 0.64 & 0.66 & 0.78 & 0.75 &  &  &  \\
\qquad Semantic entropy & 0.72 & 0.58 & 0.61 & 0.69 & 0.82 & 0.85 &  &  \\
\qquad Semantic sets conf. & 0.72 & 0.59 & 0.61 & 0.69 & 0.81 & 0.85 &  &  \\
\qquad Cosine similarity & 0.74 & 0.61 & 0.66 & 0.82 & 0.79 & 0.76 &  &  \\
\qquad Equivalence rate &  &  &  &  &  & 0.87 &  &  \\
\qquad CodeBLEU consistency &  &  &  &  &  & 0.80 &  &  \\
\multicolumn{9}{l}{\textit{Consistency-based white-box}} \\
\qquad CoCoA & 0.77 & 0.68 & 0.70 & 0.81 & 0.80 & 0.79 &  &  \\
\qquad Monte Carlo prob. & 0.77 & 0.75 & 0.66 & 0.77 & 0.68 & 0.71 &  &  \\
\qquad WB semantic entropy & 0.53 & 0.51 & 0.50 & 0.50 & 0.75 & 0.85 &  &  \\
\qquad Semantic density & 0.81 & 0.61 & 0.49 & 0.48 & 0.79 &  &  &  \\
\multicolumn{9}{l}{\textit{Reflexive}} \\
\qquad Verbalized confidence & 0.73 & 0.68 & 0.54 & 0.55 & 0.59 & 0.80 & 0.61 & 0.57 \\
\qquad P(True) & \textbf{0.86} & 0.86 & 0.69 & 0.71 & 0.64 & 0.84 & 0.60 & 0.56 \\
\multicolumn{9}{l}{\textit{Graph-based}} \\
\qquad Degree centrality &  &  &  &  &  &  & 0.70 & 0.67 \\
\qquad Betweenness centrality &  &  &  &  &  &  & 0.67 & 0.64 \\
\qquad Closeness centrality &  &  &  &  &  &  & 0.68 & 0.67 \\
\qquad Harmonic centrality &  &  &  &  &  &  & 0.68 & 0.66 \\
\qquad Laplacian centrality &  &  &  &  &  &  & 0.69 & 0.66 \\
\qquad PageRank &  &  &  &  &  &  & 0.69 & 0.67 \\
\multicolumn{9}{l}{\textit{Ensembles}} \\
\qquad Ensemble (avg) & 0.84 & 0.80 & 0.73 & 0.82 & \textbf{0.83} & \textbf{0.89} & \textbf{0.72} & 0.69 \\
\qquad Logistic regression & 0.83 & 0.85 & 0.72 & \textbf{0.82} & 0.83 & 0.89 & 0.72 & \textbf{0.70} \\
\qquad Random forest & 0.85 & \textbf{0.90} & \textbf{0.74} & 0.81 & 0.82 & 0.89 & 0.70 & 0.70 \\
\qquad Gradient boosting & 0.81 & 0.88 & 0.74 & 0.79 & 0.82 & 0.87 & 0.69 & 0.68 \\
\bottomrule
\end{tabular}
\caption{Gemini-2.5-Flash AUROC performance of uncertainty quantification methods across nine benchmarks spanning short-form QA, code generation, and long-form generation domains.}
\label{tab:auroc_results_flash}
\end{table*} 

\begin{table*}[t]
\centering
\tiny
\setlength{\tabcolsep}{3pt}
\begin{tabular}{@{}lcccccccc@{}}
\toprule
 & \multicolumn{5}{c}{\textbf{Short-form QA}} & \textbf{Code} & \multicolumn{2}{c}{\textbf{Long-form}} \\
\cmidrule(lr){2-6} \cmidrule(lr){7-7} \cmidrule(lr){8-9}
\textbf{Scorer} & BigMath & OpenR1 & DROP & Hotpot & SimpleQA & LiveCodeBench & Mushroom-Fact & River-Fact \\
\midrule
\multicolumn{9}{l}{\textit{Single-generation white-box}} \\
\qquad Sequence prob. & 0.72 & 0.62 & 0.69 & 0.73 & 0.54 & 0.58 &  &  \\
\qquad Norm. sequence prob. & 0.71 & 0.59 & 0.66 & 0.73 & 0.54 &  &  &  \\
\qquad Min. token prob. & 0.72 & 0.61 & 0.69 & 0.73 & 0.54 & 0.76 &  &  \\
\qquad Probability margin & 0.71 & 0.59 & 0.66 & 0.72 & 0.54 & 0.55 &  &  \\
\qquad Mean token entropy & 0.71 & 0.59 & 0.65 & 0.73 & 0.54 & 0.58 &  &  \\
\qquad Min. token entropy & 0.72 & 0.61 & 0.69 & 0.74 & 0.54 & 0.77 &  &  \\
\multicolumn{9}{l}{\textit{Consistency-based black-box}} \\
\qquad Exact match rate & 0.77 & 0.77 & 0.65 & 0.76 & 0.81 &  &  &  \\
\qquad Non-contradiction prob. & 0.77 & 0.83 & 0.63 & 0.75 & 0.84 &  &  &  \\
\qquad Entailment prob. & 0.77 & 0.81 & 0.64 & 0.76 & 0.85 &  &  &  \\
\qquad BERTScore consistency & 0.78 & 0.72 & 0.65 & 0.77 & 0.76 &  &  &  \\
\qquad Semantic entropy & 0.74 & 0.78 & 0.57 & 0.66 & 0.83 & 0.82 &  &  \\
\qquad Semantic sets conf. & 0.74 & 0.78 & 0.57 & 0.66 & 0.83 & 0.82 &  &  \\
\qquad Cosine similarity & 0.76 & 0.73 & 0.64 & 0.76 & 0.81 & 0.80 &  &  \\
\qquad Equivalence rate &  &  &  &  &  & 0.82 &  &  \\
\qquad CodeBLEU consistency &  &  &  &  &  & 0.80 &  &  \\
\multicolumn{9}{l}{\textit{Consistency-based white-box}} \\
\qquad CoCoA & 0.82 & 0.72 & 0.66 & 0.81 & 0.82 & 0.79 &  &  \\
\qquad Monte Carlo prob. & 0.75 & 0.66 & 0.68 & 0.77 & 0.55 & 0.60 &  &  \\
\qquad WB semantic entropy & 0.52 & 0.50 & 0.50 & 0.50 & 0.76 & 0.82 &  &  \\
\qquad Semantic density & 0.79 & 0.73 & 0.51 & 0.48 & 0.84 &  &  &  \\
\multicolumn{9}{l}{\textit{Reflexive}} \\
\qquad Verbalized confidence & 0.72 & 0.68 & 0.51 & 0.58 & 0.59 & 0.72 & 0.55 & 0.54 \\
\qquad P(True) & 0.81 & 0.79 & 0.57 & 0.69 & 0.74 & 0.84 & 0.59 & 0.56 \\
\multicolumn{9}{l}{\textit{Graph-based}} \\
\qquad Degree centrality &  &  &  &  &  &  & 0.66 & 0.67 \\
\qquad Betweenness centrality &  &  &  &  &  &  & 0.63 & 0.63 \\
\qquad Closeness centrality &  &  &  &  &  &  & 0.63 & 0.66 \\
\qquad Harmonic centrality &  &  &  &  &  &  & 0.63 & 0.65 \\
\qquad Laplacian centrality &  &  &  &  &  &  & 0.64 & 0.66 \\
\qquad PageRank &  &  &  &  &  &  & 0.64 & 0.66 \\
\multicolumn{9}{l}{\textit{Ensembles}} \\
\qquad Ensemble (avg) & \textbf{0.88} & 0.86 & 0.69 & 0.84 & 0.86 & \textbf{0.85} & 0.67 & 0.68 \\
\qquad Logistic regression & 0.87 & 0.90 & 0.69 & 0.82 & \textbf{0.86} & 0.85 & \textbf{0.67} & \textbf{0.69} \\
\qquad Random forest & 0.88 & \textbf{0.90} & \textbf{0.74} & \textbf{0.84} & 0.86 & 0.85 & 0.66 & 0.69 \\
\qquad Gradient boosting & 0.87 & 0.89 & 0.72 & 0.84 & 0.86 & 0.84 & 0.65 & 0.67 \\
\bottomrule
\end{tabular}
\caption{Gemini-2.5-Pro AUROC performance of uncertainty quantification methods across nine benchmarks spanning short-form QA, code generation, and long-form generation domains.}
\label{tab:auroc_results_pro}
\end{table*} 

\begin{table*}[t]
\centering
\tiny
\setlength{\tabcolsep}{3pt}
\begin{tabular}{@{}lcccccccc@{}}
\toprule
 & \multicolumn{5}{c}{\textbf{Short-form QA}} & \textbf{Code} & \multicolumn{2}{c}{\textbf{Long-form}} \\
\cmidrule(lr){2-6} \cmidrule(lr){7-7} \cmidrule(lr){8-9}
\textbf{Scorer} & BigMath & OpenR1 & DROP & Hotpot & SimpleQA & LiveCodeBench & Mushroom-Fact & River-Fact \\
\midrule
\multicolumn{9}{l}{\textit{Single-generation white-box}} \\
\qquad Sequence prob. & 0.82 & 0.81 & 0.69 & 0.71 & 0.81 & 0.73 &  &  \\
\qquad Norm. sequence prob. & 0.84 & 0.81 & 0.62 & 0.71 & 0.83 &  &  &  \\
\qquad Min. token prob. & 0.82 & 0.81 & 0.67 & 0.70 & 0.80 & 0.71 &  &  \\
\qquad Probability margin & 0.82 & 0.76 & 0.59 & 0.71 & 0.81 & 0.75 &  &  \\
\qquad Mean token entropy & 0.84 & 0.80 & 0.61 & 0.73 & 0.84 & 0.75 &  &  \\
\qquad Min. token entropy & 0.82 & 0.80 & 0.70 & 0.73 & 0.82 & 0.76 &  &  \\
\multicolumn{9}{l}{\textit{Consistency-based black-box}} \\
\qquad Exact match rate & 0.78 & 0.78 & 0.69 & 0.67 & 0.78 &  &  &  \\
\qquad Non-contradiction prob. & 0.79 & 0.78 & 0.71 & 0.70 & 0.83 &  &  &  \\
\qquad Entailment prob. & 0.79 & 0.78 & 0.71 & 0.70 & 0.83 &  &  &  \\
\qquad BERTScore consistency & 0.78 & 0.73 & 0.63 & 0.68 & 0.72 &  &  &  \\
\qquad Semantic entropy & 0.80 & 0.78 & 0.67 & 0.67 & 0.84 & 0.85 &  &  \\
\qquad Semantic sets conf. & 0.80 & 0.78 & 0.67 & 0.67 & 0.84 & 0.85 &  &  \\
\qquad Cosine similarity & 0.78 & 0.77 & 0.59 & 0.69 & 0.78 & 0.72 &  &  \\
\qquad Equivalence rate &  &  &  &  &  & 0.86 &  &  \\
\qquad CodeBLEU consistency &  &  &  &  &  & 0.74 &  &  \\
\multicolumn{9}{l}{\textit{Consistency-based white-box}} \\
\qquad CoCoA & 0.84 & 0.81 & 0.61 & 0.72 & 0.83 & 0.74 &  &  \\
\qquad Monte Carlo prob. & 0.83 & 0.81 & 0.64 & 0.74 & 0.83 & 0.82 &  &  \\
\qquad WB semantic entropy & 0.50 & 0.51 & 0.50 & 0.50 & 0.79 & 0.78 &  &  \\
\qquad Semantic density & 0.74 & 0.73 & 0.61 & 0.52 & 0.83 &  &  &  \\
\multicolumn{9}{l}{\textit{Reflexive}} \\
\qquad Verbalized confidence & 0.59 & 0.58 & 0.58 & 0.61 & 0.69 & 0.68 & 0.61 & 0.56 \\
\qquad P(True) & 0.68 & 0.64 & 0.70 & 0.70 & 0.78 & 0.82 & 0.70 & 0.65 \\
\multicolumn{9}{l}{\textit{Graph-based}} \\
\qquad Degree centrality &  &  &  &  &  &  & 0.72 & 0.67 \\
\qquad Betweenness centrality &  &  &  &  &  &  & 0.70 & 0.64 \\
\qquad Closeness centrality &  &  &  &  &  &  & 0.72 & 0.67 \\
\qquad Harmonic centrality &  &  &  &  &  &  & 0.72 & 0.66 \\
\qquad Laplacian centrality &  &  &  &  &  &  & 0.72 & 0.64 \\
\qquad PageRank &  &  &  &  &  &  & 0.71 & 0.64 \\
\multicolumn{9}{l}{\textit{Ensembles}} \\
\qquad Ensemble (avg) & 0.84 & 0.81 & 0.76 & 0.75 & 0.87 & \textbf{0.88} & 0.74 & 0.68 \\
\qquad Logistic regression & 0.84 & 0.81 & \textbf{0.77} & \textbf{0.77} & \textbf{0.87} & 0.88 & \textbf{0.74} & 0.68 \\
\qquad Random forest & \textbf{0.85} & \textbf{0.82} & 0.77 & 0.74 & 0.87 & 0.88 & 0.74 & \textbf{0.71} \\
\qquad Gradient boosting & 0.85 & 0.81 & 0.75 & 0.69 & 0.86 & 0.87 & 0.74 & 0.70 \\
\bottomrule
\end{tabular}
\caption{GPT-4o AUROC performance of uncertainty quantification methods across nine benchmarks spanning short-form QA, code generation, and long-form generation domains.}
\label{tab:auroc_results_4o}
\end{table*} 

\begin{table*}[t]
\centering
\tiny
\setlength{\tabcolsep}{3pt}
\begin{tabular}{@{}lcccccccc@{}}
\toprule
 & \multicolumn{5}{c}{\textbf{Short-form QA}} & \textbf{Code} & \multicolumn{2}{c}{\textbf{Long-form}} \\
\cmidrule(lr){2-6} \cmidrule(lr){7-7} \cmidrule(lr){8-9}
\textbf{Scorer} & BigMath & OpenR1 & DROP & Hotpot & SimpleQA & LiveCodeBench & Mushroom-Fact & River-Fact \\
\midrule
\multicolumn{9}{l}{\textit{Single-generation white-box}} \\
\qquad Sequence prob. & 0.53 & 0.66 & 0.63 & 0.74 & 0.74 & 0.76 &  &  \\
\qquad Norm. sequence prob. & 0.79 & 0.71 & 0.60 & 0.73 & 0.69 &  &  &  \\
\qquad Min. token prob. & 0.55 & 0.67 & 0.62 & 0.73 & 0.71 & 0.74 &  &  \\
\qquad Probability margin & 0.82 & 0.68 & 0.59 & 0.74 & 0.66 & 0.75 &  &  \\
\qquad Mean token entropy & 0.85 & 0.71 & 0.60 & 0.75 & 0.72 & 0.78 &  &  \\
\qquad Min. token entropy & 0.59 & 0.69 & 0.63 & 0.74 & \textbf{0.78} & 0.78 &  &  \\
\multicolumn{9}{l}{\textit{Consistency-based black-box}} \\
\qquad Exact match rate & 0.53 & 0.65 & 0.60 & 0.74 & 0.73 &  &  &  \\
\qquad Non-contradiction prob. & 0.74 & 0.72 & 0.64 & 0.73 & 0.75 &  &  &  \\
\qquad Entailment prob. & 0.72 & 0.72 & 0.64 & 0.75 & 0.76 &  &  &  \\
\qquad BERTScore consistency & 0.52 & 0.63 & 0.57 & 0.71 & 0.66 &  &  &  \\
\qquad Semantic entropy & 0.74 & 0.74 & 0.64 & 0.65 & 0.77 & 0.86 &  &  \\
\qquad Semantic sets conf. & 0.74 & 0.74 & 0.64 & 0.65 & 0.78 & 0.85 &  &  \\
\qquad Cosine similarity & 0.62 & 0.69 & 0.55 & 0.70 & 0.69 & 0.74 &  &  \\
\qquad Equivalence rate &  &  &  &  &  & 0.86 &  &  \\
\qquad CodeBLEU consistency &  &  &  &  &  & 0.76 &  &  \\
\multicolumn{9}{l}{\textit{Consistency-based white-box}} \\
\qquad CoCoA & 0.77 & 0.72 & 0.58 & 0.73 & 0.70 & 0.77 &  &  \\
\qquad Monte Carlo prob. & 0.76 & 0.73 & 0.58 & 0.76 & 0.74 & 0.81 &  &  \\
\qquad WB semantic entropy & 0.54 & 0.54 & 0.50 & 0.50 & 0.71 & 0.85 &  &  \\
\qquad Semantic density & 0.60 & 0.64 & 0.59 & 0.50 & 0.71 &  &  &  \\
\multicolumn{9}{l}{\textit{Reflexive}} \\
\qquad Verbalized confidence & 0.70 & 0.61 & 0.59 & 0.63 & 0.57 & 0.72 & 0.60 & 0.57 \\
\qquad P(True) & 0.78 & 0.69 & 0.65 & 0.69 & 0.65 & 0.76 & 0.66 & 0.59 \\
\multicolumn{9}{l}{\textit{Graph-based}} \\
\qquad Degree centrality &  &  &  &  &  &  & 0.71 & 0.66 \\
\qquad Betweenness centrality &  &  &  &  &  &  & 0.68 & 0.64 \\
\qquad Closeness centrality &  &  &  &  &  &  & 0.71 & 0.66 \\
\qquad Harmonic centrality &  &  &  &  &  &  & 0.71 & 0.65 \\
\qquad Laplacian centrality &  &  &  &  &  &  & 0.70 & 0.64 \\
\qquad PageRank &  &  &  &  &  &  & 0.71 & 0.64 \\
\multicolumn{9}{l}{\textit{Ensembles}} \\
\qquad Ensemble (avg) & 0.85 & 0.77 & 0.67 & \textbf{0.78} & 0.78 & \textbf{0.88} & \textbf{0.72} & 0.67 \\
\qquad Logistic regression & 0.87 & 0.77 & 0.67 & 0.78 & 0.78 & 0.88 & 0.72 & 0.68 \\
\qquad Random forest & \textbf{0.88} & \textbf{0.78} & \textbf{0.70} & 0.78 & 0.77 & 0.88 & 0.71 & \textbf{0.69} \\
\qquad Gradient boosting & 0.88 & 0.77 & 0.69 & 0.77 & 0.76 & 0.88 & 0.70 & 0.66 \\
\bottomrule
\end{tabular}
\caption{GPT-4o-mini AUROC performance of uncertainty quantification methods across nine benchmarks spanning short-form QA, code generation, and long-form generation domains.}
\label{tab:auroc_results_mini}
\end{table*} 

\begin{table*}[t]
\centering
\tiny
\setlength{\tabcolsep}{3pt}
\begin{tabular}{@{}lcccccccc@{}}
\toprule
 & \multicolumn{5}{c}{\textbf{Short-form QA}} & \textbf{Code} & \multicolumn{2}{c}{\textbf{Long-form}} \\
\cmidrule(lr){2-6} \cmidrule(lr){7-7} \cmidrule(lr){8-9}
\textbf{Scorer} & BigMath & OpenR1 & DROP & Hotpot & SimpleQA & LiveCodeBench & Mushroom-Fact & River-Fact \\
\midrule
\multicolumn{9}{l}{\textit{Single-generation white-box}} \\
\qquad Sequence prob. & 0.21 & 0.58 & 0.14 & 0.07 & 0.55 & 0.05 &  &  \\
\qquad Norm. sequence prob. & 0.04 & 0.04 & 0.15 & 0.07 & 0.62 &  &  &  \\
\qquad Min. token prob. & 0.21 & 0.56 & 0.14 & 0.07 & 0.56 & 0.73 &  &  \\
\qquad Probability margin & 0.05 & 0.05 & 0.15 & 0.07 & 0.62 & 0.06 &  &  \\
\qquad Mean token entropy & 0.05 & 0.07 & 0.15 & 0.06 & 0.64 & 0.09 &  &  \\
\qquad Min. token entropy & 0.15 & 0.44 & 0.13 & 0.05 & 0.58 & 0.53 &  &  \\
\multicolumn{9}{l}{\textit{Consistency-based black-box}} \\
\qquad Exact match rate & 0.22 & 0.58 & 0.12 & 0.07 & 0.14 &  &  &  \\
\qquad Non-contradiction prob. & 0.03 & 0.04 & 0.14 & 0.06 & 0.23 &  &  &  \\
\qquad Entailment prob. & 0.07 & 0.14 & 0.11 & 0.04 & 0.18 &  &  &  \\
\qquad BERTScore consistency & \textbf{0.02} & \textbf{0.02} & 0.14 & 0.05 & 0.63 &  &  &  \\
\qquad Semantic entropy & 0.08 & 0.13 & 0.13 & 0.05 & 0.19 & 0.08 &  &  \\
\qquad Semantic sets conf. & 0.06 & 0.11 & 0.14 & 0.05 & 0.21 & 0.04 &  &  \\
\qquad Cosine similarity & 0.03 & 0.05 & 0.13 & 0.05 & 0.56 & 0.05 &  &  \\
\qquad Equivalence rate &  &  &  &  &  & 0.11 &  &  \\
\qquad CodeBLEU consistency &  &  &  &  &  & 0.43 &  &  \\
\multicolumn{9}{l}{\textit{Consistency-based white-box}} \\
\qquad CoCoA & 0.05 & 0.07 & 0.12 & 0.05 & 0.49 & 0.07 &  &  \\
\qquad Monte Carlo prob. & 0.03 & 0.04 & 0.14 & 0.05 & 0.62 & 0.05 &  &  \\
\qquad WB semantic entropy & 0.06 & 0.10 & 0.16 & 0.07 & 0.32 & 0.08 &  &  \\
\qquad Semantic density & 0.03 & 0.04 & 0.14 & 0.04 & 0.29 &  &  &  \\
\multicolumn{9}{l}{\textit{Reflexive}} \\
\qquad Verbalized confidence & 0.04 & 0.06 & 0.16 & 0.08 & 0.54 & 0.07 & 0.37 & 0.45 \\
\qquad P(True) & 0.05 & 0.07 & 0.17 & 0.07 & 0.52 & 0.11 & 0.42 & 0.47 \\
\multicolumn{9}{l}{\textit{Graph-based}} \\
\qquad Degree centrality &  &  &  &  &  &  & 0.14 & 0.18 \\
\qquad Betweenness centrality &  &  &  &  &  &  & 0.54 & 0.49 \\
\qquad Closeness centrality &  &  &  &  &  &  & 0.34 & 0.36 \\
\qquad Harmonic centrality &  &  &  &  &  &  & 0.35 & 0.38 \\
\qquad Laplacian centrality &  &  &  &  &  &  & 0.52 & 0.46 \\
\qquad PageRank &  &  &  &  &  &  & 0.54 & 0.49 \\
\multicolumn{9}{l}{\textit{Ensembles}} \\
\qquad Ensemble (avg) & 0.03 & 0.10 & 0.12 & 0.04 & 0.42 & 0.11 & 0.13 & 0.27 \\
\qquad Logistic regression & 0.02 & 0.03 & \textbf{0.04} & \textbf{0.02} & 0.08 & 0.04 & \textbf{0.05} & 0.03 \\
\qquad Random forest & 0.02 & 0.03 & 0.04 & 0.02 & \textbf{0.05} & \textbf{0.03} & 0.06 & \textbf{0.03} \\
\qquad Gradient boosting & 0.03 & 0.06 & 0.04 & 0.04 & 0.08 & 0.06 & 0.07 & 0.07 \\
\bottomrule
\end{tabular}
\caption{Gemini-2.5-Flash ECE of uncertainty quantification methods across nine benchmarks spanning short-form QA, code generation, and long-form generation domains.}
\label{tab:ece_results_flash}
\end{table*} 

\begin{table*}[t]
\centering
\tiny
\setlength{\tabcolsep}{3pt}
\begin{tabular}{@{}lcccccccc@{}}
\toprule
 & \multicolumn{5}{c}{\textbf{Short-form QA}} & \textbf{Code} & \multicolumn{2}{c}{\textbf{Long-form}} \\
\cmidrule(lr){2-6} \cmidrule(lr){7-7} \cmidrule(lr){8-9}
\textbf{Scorer} & BigMath & OpenR1 & DROP & Hotpot & SimpleQA & LiveCodeBench & Mushroom-Fact & River-Fact \\
\midrule
\multicolumn{9}{l}{\textit{Single-generation white-box}} \\
\qquad Sequence prob. & 0.08 & 0.16 & 0.15 & 0.07 & 0.46 & 0.02 &  &  \\
\qquad Norm. sequence prob. & 0.07 & 0.07 & 0.15 & 0.06 & 0.46 &  &  &  \\
\qquad Min. token prob. & 0.08 & 0.16 & 0.15 & 0.06 & 0.46 & 0.82 &  &  \\
\qquad Probability margin & 0.07 & 0.08 & 0.15 & 0.06 & 0.46 & 0.02 &  &  \\
\qquad Mean token entropy & 0.07 & 0.07 & 0.15 & 0.06 & 0.46 & 0.05 &  &  \\
\qquad Min. token entropy & 0.07 & 0.13 & 0.14 & 0.05 & 0.46 & 0.54 &  &  \\
\multicolumn{9}{l}{\textit{Consistency-based black-box}} \\
\qquad Exact match rate & 0.05 & 0.12 & 0.12 & 0.08 & 0.11 &  &  &  \\
\qquad Non-contradiction prob. & 0.04 & 0.04 & 0.15 & 0.05 & 0.20 &  &  &  \\
\qquad Entailment prob. & 0.04 & 0.06 & 0.13 & 0.04 & 0.16 &  &  &  \\
\qquad BERTScore consistency & 0.06 & 0.07 & 0.14 & 0.04 & 0.42 &  &  &  \\
\qquad Semantic entropy & 0.04 & 0.06 & 0.14 & 0.05 & 0.17 & 0.15 &  &  \\
\qquad Semantic sets conf. & 0.04 & 0.04 & 0.14 & 0.05 & 0.17 & 0.08 &  &  \\
\qquad Cosine similarity & 0.05 & 0.05 & 0.13 & 0.04 & 0.37 & 0.04 &  &  \\
\qquad Equivalence rate &  &  &  &  &  & 0.20 &  &  \\
\qquad CodeBLEU consistency &  &  &  &  &  & 0.47 &  &  \\
\multicolumn{9}{l}{\textit{Consistency-based white-box}} \\
\qquad CoCoA & 0.05 & 0.07 & 0.13 & 0.05 & 0.37 & 0.08 &  &  \\
\qquad Monte Carlo prob. & 0.06 & 0.06 & 0.15 & 0.05 & 0.46 & \textbf{0.02} &  &  \\
\qquad WB semantic entropy & 0.07 & 0.08 & 0.16 & 0.06 & 0.25 & 0.15 &  &  \\
\qquad Semantic density & 0.05 & 0.06 & 0.13 & 0.03 & 0.22 &  &  &  \\
\multicolumn{9}{l}{\textit{Reflexive}} \\
\qquad Verbalized confidence & 0.05 & 0.07 & 0.16 & 0.07 & 0.38 & 0.06 & 0.41 & 0.48 \\
\qquad P(True) & 0.04 & 0.06 & 0.17 & 0.07 & 0.34 & 0.08 & 0.41 & 0.49 \\
\multicolumn{9}{l}{\textit{Graph-based}} \\
\qquad Degree centrality &  &  &  &  &  &  & 0.19 & 0.20 \\
\qquad Betweenness centrality &  &  &  &  &  &  & 0.56 & 0.47 \\
\qquad Closeness centrality &  &  &  &  &  &  & 0.34 & 0.39 \\
\qquad Harmonic centrality &  &  &  &  &  &  & 0.35 & 0.40 \\
\qquad Laplacian centrality &  &  &  &  &  &  & 0.53 & 0.44 \\
\qquad PageRank &  &  &  &  &  &  & 0.55 & 0.47 \\
\multicolumn{9}{l}{\textit{Ensembles}} \\
\qquad Ensemble (avg) & 0.04 & 0.04 & 0.13 & 0.04 & 0.31 & 0.19 & 0.16 & 0.18 \\
\qquad Logistic regression & 0.03 & \textbf{0.03} & \textbf{0.03} & \textbf{0.01} & \textbf{0.05} & 0.03 & 0.06 & 0.03 \\
\qquad Random forest & \textbf{0.02} & 0.03 & 0.04 & 0.02 & 0.06 & 0.03 & \textbf{0.06} & \textbf{0.03} \\
\qquad Gradient boosting & 0.04 & 0.05 & 0.05 & 0.03 & 0.13 & 0.04 & 0.08 & 0.06 \\
\bottomrule
\end{tabular}
\caption{Gemini-2.5-Pro ECE of uncertainty quantification methods across nine benchmarks spanning short-form QA, code generation, and long-form generation domains.}
\label{tab:ece_results_pro}
\end{table*} 

\begin{table*}[t]
\centering
\tiny
\setlength{\tabcolsep}{3pt}
\begin{tabular}{@{}lcccccccc@{}}
\toprule
 & \multicolumn{5}{c}{\textbf{Short-form QA}} & \textbf{Code} & \multicolumn{2}{c}{\textbf{Long-form}} \\
\cmidrule(lr){2-6} \cmidrule(lr){7-7} \cmidrule(lr){8-9}
\textbf{Scorer} & BigMath & OpenR1 & DROP & Hotpot & SimpleQA & LiveCodeBench & Mushroom-Fact & River-Fact \\
\midrule
\multicolumn{9}{l}{\textit{Single-generation white-box}} \\
\qquad Sequence prob. & 0.19 & 0.17 & 0.34 & 0.21 & 0.14 & 0.22 &  &  \\
\qquad Norm. sequence prob. & 0.20 & 0.19 & 0.13 & 0.05 & 0.34 &  &  &  \\
\qquad Min. token prob. & 0.19 & 0.17 & 0.29 & 0.19 & 0.14 & 0.50 &  &  \\
\qquad Probability margin & 0.18 & 0.19 & 0.13 & 0.04 & 0.43 & 0.32 &  &  \\
\qquad Mean token entropy & 0.26 & 0.30 & 0.15 & 0.03 & 0.50 & 0.36 &  &  \\
\qquad Min. token entropy & 0.25 & 0.26 & 0.08 & 0.09 & 0.25 & 0.25 &  &  \\
\multicolumn{9}{l}{\textit{Consistency-based black-box}} \\
\qquad Exact match rate & 0.24 & 0.25 & 0.33 & 0.22 & 0.13 &  &  &  \\
\qquad Non-contradiction prob. & 0.26 & 0.29 & 0.17 & 0.06 & 0.29 &  &  &  \\
\qquad Entailment prob. & 0.24 & 0.26 & 0.07 & 0.08 & 0.23 &  &  &  \\
\qquad BERTScore consistency & 0.58 & 0.72 & 0.17 & 0.04 & 0.66 &  &  &  \\
\qquad Semantic entropy & 0.24 & 0.26 & 0.16 & 0.05 & 0.24 & 0.06 &  &  \\
\qquad Semantic sets conf. & 0.26 & 0.28 & 0.17 & 0.05 & 0.24 & 0.10 &  &  \\
\qquad Cosine similarity & 0.52 & 0.65 & 0.15 & 0.04 & 0.58 & 0.36 &  &  \\
\qquad Equivalence rate &  &  &  &  &  & 0.13 &  &  \\
\qquad CodeBLEU consistency &  &  &  &  &  & 0.11 &  &  \\
\multicolumn{9}{l}{\textit{Consistency-based white-box}} \\
\qquad CoCoA & 0.17 & 0.17 & 0.12 & 0.08 & 0.27 & 0.17 &  &  \\
\qquad Monte Carlo prob. & 0.18 & 0.19 & 0.08 & 0.03 & 0.35 & 0.11 &  &  \\
\qquad WB semantic entropy & 0.60 & 0.73 & 0.22 & 0.07 & 0.41 & 0.11 &  &  \\
\qquad Semantic density & 0.56 & 0.68 & 0.18 & 0.04 & 0.29 &  &  &  \\
\multicolumn{9}{l}{\textit{Reflexive}} \\
\qquad Verbalized confidence & 0.52 & 0.58 & 0.19 & 0.06 & 0.49 & 0.30 & 0.31 & 0.40 \\
\qquad P(True) & 0.20 & 0.18 & 0.17 & 0.16 & 0.18 & 0.30 & 0.26 & 0.37 \\
\multicolumn{9}{l}{\textit{Graph-based}} \\
\qquad Degree centrality &  &  &  &  &  &  & 0.13 & 0.16 \\
\qquad Betweenness centrality &  &  &  &  &  &  & 0.60 & 0.53 \\
\qquad Closeness centrality &  &  &  &  &  &  & 0.23 & 0.31 \\
\qquad Harmonic centrality &  &  &  &  &  &  & 0.24 & 0.32 \\
\qquad Laplacian centrality &  &  &  &  &  &  & 0.57 & 0.50 \\
\qquad PageRank &  &  &  &  &  &  & 0.60 & 0.53 \\
\multicolumn{9}{l}{\textit{Ensembles}} \\
\qquad Ensemble (avg) & 0.30 & 0.31 & 0.08 & 0.03 & 0.32 & 0.17 & 0.11 & 0.17 \\
\qquad Logistic regression & 0.10 & 0.08 & \textbf{0.04} & 0.02 & \textbf{0.05} & 0.05 & 0.07 & 0.03 \\
\qquad Random forest & \textbf{0.06} & \textbf{0.05} & 0.04 & \textbf{0.02} & 0.05 & \textbf{0.05} & \textbf{0.05} & \textbf{0.03} \\
\qquad Gradient boosting & 0.07 & 0.08 & 0.07 & 0.04 & 0.08 & 0.08 & 0.06 & 0.05 \\
\bottomrule
\end{tabular}
\caption{GPT-4o ECE of uncertainty quantification methods across nine benchmarks spanning short-form QA, code generation, and long-form generation domains.}
\label{tab:ece_results_4o}
\end{table*} 

\begin{table*}[t]
\centering
\tiny
\setlength{\tabcolsep}{3pt}
\begin{tabular}{@{}lcccccccc@{}}
\toprule
 & \multicolumn{5}{c}{\textbf{Short-form QA}} & \textbf{Code} & \multicolumn{2}{c}{\textbf{Long-form}} \\
\cmidrule(lr){2-6} \cmidrule(lr){7-7} \cmidrule(lr){8-9}
\textbf{Scorer} & BigMath & OpenR1 & DROP & Hotpot & SimpleQA & LiveCodeBench & Mushroom-Fact & River-Fact \\
\midrule
\multicolumn{9}{l}{\textit{Single-generation white-box}} \\
\qquad Sequence prob. & 0.35 & 0.21 & 0.30 & 0.24 & 0.15 & 0.35 &  &  \\
\qquad Norm. sequence prob. & 0.23 & 0.27 & 0.17 & 0.06 & 0.45 &  &  &  \\
\qquad Min. token prob. & 0.35 & 0.21 & 0.25 & 0.20 & 0.19 & 0.39 &  &  \\
\qquad Probability margin & 0.25 & 0.33 & 0.17 & 0.06 & 0.59 & 0.41 &  &  \\
\qquad Mean token entropy & 0.31 & 0.42 & 0.19 & 0.08 & 0.67 & 0.44 &  &  \\
\qquad Min. token entropy & 0.26 & 0.26 & 0.07 & 0.08 & 0.31 & 0.09 &  &  \\
\multicolumn{9}{l}{\textit{Consistency-based black-box}} \\
\qquad Exact match rate & 0.36 & 0.26 & 0.28 & 0.22 & 0.19 &  &  &  \\
\qquad Non-contradiction prob. & 0.33 & 0.39 & 0.21 & 0.09 & 0.34 &  &  &  \\
\qquad Entailment prob. & 0.24 & 0.30 & 0.11 & 0.07 & 0.27 &  &  &  \\
\qquad BERTScore consistency & 0.49 & 0.69 & 0.21 & 0.09 & 0.85 &  &  &  \\
\qquad Semantic entropy & 0.27 & 0.32 & 0.20 & 0.09 & 0.29 & 0.07 &  &  \\
\qquad Semantic sets conf. & 0.29 & 0.33 & 0.20 & 0.09 & 0.28 & 0.12 &  &  \\
\qquad Cosine similarity & 0.44 & 0.62 & 0.20 & 0.07 & 0.77 & 0.43 &  &  \\
\qquad Equivalence rate &  &  &  &  &  & 0.11 &  &  \\
\qquad CodeBLEU consistency &  &  &  &  &  & 0.09 &  &  \\
\multicolumn{9}{l}{\textit{Consistency-based white-box}} \\
\qquad CoCoA & 0.18 & 0.23 & 0.15 & 0.05 & 0.39 & 0.31 &  &  \\
\qquad Monte Carlo prob. & 0.19 & 0.23 & 0.14 & 0.03 & 0.45 & 0.32 &  &  \\
\qquad WB semantic entropy & 0.51 & 0.70 & 0.25 & 0.11 & 0.48 & 0.07 &  &  \\
\qquad Semantic density & 0.46 & 0.63 & 0.21 & 0.08 & 0.36 &  &  &  \\
\multicolumn{9}{l}{\textit{Reflexive}} \\
\qquad Verbalized confidence & 0.23 & 0.35 & 0.12 & 0.10 & 0.59 & 0.28 & 0.26 & 0.33 \\
\qquad P(True) & 0.25 & 0.27 & 0.25 & 0.15 & 0.26 & 0.35 & 0.43 & 0.48 \\
\multicolumn{9}{l}{\textit{Graph-based}} \\
\qquad Degree centrality &  &  &  &  &  &  & 0.12 & 0.18 \\
\qquad Betweenness centrality &  &  &  &  &  &  & 0.50 & 0.46 \\
\qquad Closeness centrality &  &  &  &  &  &  & 0.32 & 0.38 \\
\qquad Harmonic centrality &  &  &  &  &  &  & 0.35 & 0.40 \\
\qquad Laplacian centrality &  &  &  &  &  &  & 0.47 & 0.43 \\
\qquad PageRank &  &  &  &  &  &  & 0.50 & 0.46 \\
\multicolumn{9}{l}{\textit{Ensembles}} \\
\qquad Ensemble (avg) & 0.25 & 0.33 & 0.13 & 0.04 & 0.41 & 0.21 & 0.17 & 0.23 \\
\qquad Logistic regression & 0.08 & 0.05 & 0.05 & 0.04 & 0.03 & 0.07 & 0.06 & 0.03 \\
\qquad Random forest & \textbf{0.06} & \textbf{0.05} & \textbf{0.05} & \textbf{0.03} & \textbf{0.02} & \textbf{0.06} & \textbf{0.06} & \textbf{0.03} \\
\qquad Gradient boosting & 0.10 & 0.06 & 0.11 & 0.04 & 0.03 & 0.10 & 0.06 & 0.07 \\
\bottomrule
\end{tabular}
\caption{GPT-4o-mini ECE of uncertainty quantification methods across nine benchmarks spanning short-form QA, code generation, and long-form generation domains.}
\label{tab:ece_results_mini}
\end{table*} 
\end{document}